\documentclass[a4paper, 11pt]{article}

\usepackage{amsmath}
\usepackage{amsfonts}
\usepackage{amssymb}
\usepackage{commath}

\usepackage[round]{natbib}
\usepackage{dirtytalk}
\usepackage{hyperref}

\usepackage{graphicx}
\usepackage{tikz}
\usetikzlibrary{arrows,shapes,positioning,shadows,trees,automata}
\tikzset{
  basic/.style  = {draw, text width=2cm, drop shadow, font=\sffamily, rectangle},
  root/.style   = {basic, rounded corners=2pt, thin, align=center,
                   fill=white},
  level 2/.style = {basic, rounded corners=6pt, thin,align=center, fill=white,
                   text width=8em},
  level 3/.style = {basic, thin, align=left, fill=white, text width=10em},  
  level 4/.style = {basic, thin, align=left, fill=white, text width=10em},
}
\usepackage{graphicx}
\graphicspath{{./images/}}
\usepackage{subcaption}

\usepackage[a4paper]{geometry}

\usepackage{booktabs}
\usepackage{multirow}

\usepackage{algorithm}
\usepackage{algpseudocode}

\usepackage{listings} 
\providecommand{\keywords}[1]{\\\vskip.25\baselineskip\noindent\textbf{\textit{Keywords 
			---}} #1}

\def\mycitep{\citep}
\def\mycitet{\citet}

\begin{document}

\title{ptype: Probabilistic Type Inference}

\author{Taha Ceritli\\
              School of Informatics \\
              University of Edinburgh, UK\\
              Alan Turing Institute, London, UK\\
              \texttt{t.y.ceritli@sms.ed.ac.uk}
           \and
           Christopher K.~I.~ Williams\\
           School of Informatics\\
           University of Edinburgh, UK\\
           Alan Turing Institute, London, UK\\
           \texttt{ckiw@inf.ed.ac.uk}
           \and
           James Geddes\\
            Alan Turing Institute, London, UK\\
            \texttt{jgeddes@turing.ac.uk}
}

\date{\today}

\maketitle

\begin{abstract}
Type inference refers to the task of inferring the data type of a given column of data. Current approaches often fail when data contains missing data and anomalies, which are found commonly in real-world data sets. In this paper, we propose \emph{ptype}, a probabilistic robust type inference method that allows us to detect such entries, and infer data types. We further show that the proposed method outperforms the existing methods. 
\end{abstract}

\section{Introduction}
\label{sec:introduction}
\makeatletter{\renewcommand*{\@makefnmark}{}
\footnotetext{This is a post-peer-review, pre-copyedit version of an article published in Data Mining and Knowledge Discovery. The final authenticated version is available online at: \url{http://dx.doi.org/10.1007/s10618-020-00680-1}.}\makeatother}

Data analytics can be defined as the process of generating useful
insights from raw data sets. Central to the entire process is the
concept of data wrangling, which refers to the task of understanding,
interpreting, and preparing a raw data set, and turning it into a
usable format. This often leads to a frustrating and time-consuming
process for large data sets, and even possibly for some small sized ones
\mycitep{dasu2003exploratory}. In order to design corresponding
transformations easily, one often desires to know characteristics of
given data, e.g., data types, missing data, anomalies, etc. However,
raw data often do not contain any well-documented prior information
such as meta-data.

A data column can be ascribed to a data type such as integer, string,
date, float, etc. Our focus here is on inferring the data type for each column in a table of data. Numerous
studies have attempted to tackle type inference, including wrangling
tools
\mycitep{Raman2001,Kandel2011,Guo2011,Trifacta2018,fisher2005pads,fisher2008dirt},
software packages
\mycitep{fsharp-data-pldi2016,messytables2017,tdda2018,hypoparsrcode,readr2017}, and
probabilistic approaches
\mycitep{pmlr-v70-valera17a,Vergarietal19,limaye2010annotating}. However,
often they do not work very well in the presence of missing and
anomalous data, which are commonly found in raw data sets due to the
lack of a well-organized data collection procedure.

One can develop a more accurate system by detecting missing and
anomalous entries, and performing type inference on the valid data values only. However, distinguishing such entries can be challenging
when missing data is not encoded explicitly, but as e.g., \texttt{-1} or
\texttt{-99} in a column of integer type. \mycitet{pearson2006problem} refers to such problems as the problem of
\emph{disguised missing data}. He further shows how interpreting
missing data as valid values can mislead the analysis. Similar
problems are also discussed by \mycitet{qahtan2018fahes}. Existing type inference methods do
not consider such problems comprehensively. For example, they may fail
when a column of integers contains string-valued missing data
encodings such as \texttt{null}, \texttt{NA}, etc. In this work, we
incorporate such known missing data encodings into our probabilistic
model. 

\mycitet{chandola2009anomaly} define an anomaly as
\say{a pattern that does not conform to expected normal
behavior}. In our context, anomalies refer to unexpected or invalid entries for a
given column, which might result from the data collection procedure,
e.g., error and warning messages generated by servers or the use of
open-ended entries while collecting data. The challenge here is mostly
due to the difficulty of modeling normal and anomalous data values
such that the corresponding entries become distinguishable. Moreover,
one may need to consider separate strategies to model anomalies for
different data types since their structures may vary w.r.t.\ the data
types. For example, anomalies in date columns might have a different
form than those in integer columns.

Up to now, too little attention has been paid to the aforementioned
problems by the data mining community. To this end, we introduce
ptype, a probabilistic type inference method that can robustly
annotate a column of data. The proposed model is built upon
Probabilistic Finite-State Machines (PFSMs) that are used to model
known data types, missing and anomalous data. In contrast to the
standard use of regular expressions, PFSMs have the advantage 
of generating weighted posterior
predictions even when a column of data is consistent with more than
one type model. Our method generally outperforms existing type
inference approaches for inferring data types, and also allows us to
identify missing and anomalous data entries.

In the rest of this paper, we first describe PFSMs and introduce our model (Section \ref{sec:method}). We then present the related work (Section \ref{sec:type_inference_related_work}), which is followed by the experiments and the results (Section \ref{sec:experiments}). Finally, we summarize our work and discuss the possible future research directions (Section \ref{sec:summary}).

\section{Methodology}
\label{sec:method}

This section consists of four parts. Sec.\ \ref{sec:pfsm} gives background information on PFSMs used to model regular data types, missing data, and anomalies. Sec.\ \ref{sec:model} introduces our model that uses a mixture of PFSMs. Lastly, Sec.\ \ref{sec:inference} and Sec.\ \ref{sec:training} describe respectively inference in and training of this model.

The data type, missing data and, anomalies can be defined in broad
terms as follows: The data type is the common characteristic that is
expected to be shared by entries in a column, such as integers,
strings, IP addresses, dates, etc., while missing data denotes an
absence of a data value which can be encoded in various ways,
and anomalies refer to values whose types differ from the given column
type or the missing type.

In order to model above types, we have developed PFSMs that can
generate values from the corresponding domains. This, in turn, allows
us to calculate the probability of a given data value being
generated by a particular PFSM. We then combine these PFSMs in our
model such that a data column \textbf{x} can be annotated via
probabilistic inference in the proposed model, i.e., given a column of
data, we can infer column type, and rows with missing and anomalous
values. 

\subsection{Probabilistic Finite-State Machines
  \label{sec:pfsm}} Finite-State Machines (FSMs) are a class of
mathematical models used to represent systems consisting of a finite
number of internal states. The idea is to model a system by defining
its states (including initial and final states), transitions between
the states, and external inputs/outputs. FSMs have a long history
going back at least to \mycitet{Rabin1959} and \mycitet{Gill1962}. A more recent overview of FSMs is given by \mycitet{Hopcroft2001}.

In this study, we are interested in a special type of FSMs called Probabilistic Finite-State Machines (PFSMs) in which transitions between states occur w.r.t.\ probability distributions \mycitep{Paz1971,Rabin1963}. \mycitet{Vidal2005} discuss various PFSMs in detail. Following a similar notation, we define a PFSM as a tuple $A=\left( \theta, \Sigma, \delta, I, F, T\right)$, where $\theta$ is a finite set of states, $\Sigma$ is a set of observed symbols, $\delta \subseteq \theta \times \Sigma \times \theta$ is a set of transitions among states w.r.t.\ to observed symbols, $I : \theta \rightarrow \mathbb{R}^+$ is the initial-state probabilities, $F : \theta \rightarrow \mathbb{R}^+$ is the final-state probabilities, and $T : \delta \rightarrow \mathbb{R}^+$ is the transition probabilities for elements of $\delta$. During each possible transition between states, a symbol is emitted. We denote such an event by a triple $(q,\alpha,q')$, which corresponds to a transition from a state $q \in \theta$ to a state $q' \in \theta$ emitting a symbol $\alpha \in \Sigma$. Note that $\delta$ and $T$ store respectively all the possible triples and their corresponding probabilities. 

A PFSM has to satisfy certain conditions. First, the sum of the
initial-state probabilities has to be equal to 1. Secondly, at each
state $q$, it can either transition to another state $q' \in \theta$
and emit a symbol $\alpha \in \Sigma$, or stop at state $q$ without
emitting any symbol. This can be expressed mathematically as $F(q) +
\sum_{\alpha \in \Sigma, q' \in \theta} T(q,\alpha,q')=1$ for each
state $q$, where $T(q,\alpha,q')$ represents the probability of a
triple $(q,\alpha,q')$, and $F(q)$ denotes the final-state probability
of state $q$. Based on the definition given above, a PFSM can generate
a set of strings, denoted by $\Sigma^*$\footnote{$\Sigma^*$ denotes the set of strings a PFSM can generate by emitting multiple symbols.}. For each string $s \in
\Sigma^*$, we can calculate a probability that represents how likely
it is for a given PFSM to generate the corresponding string.

Note that PFSMs resemble Hidden Markov Models (HMMs) except that we now have the final state probabilities. Recall that each state in a HMM has a probability distribution over the possible states that it can transition to. In PFSMs, each state also takes into account the probability of being a final state. Hence, the probability distribution is not only defined over the possible transitions to next states; it also includes the case of the current state being a final state. On the other hand, emissions are carried out similarly in PFSMs and HMMs: the observations are generated conditioned on the hidden states in HMMs; an observation is emitted through a transition in PFSMs since each transition is associated with a symbol. A detailed analysis of the link between PFSMs and HMMs can be found in \cite{Dupont2005}.

One can develop PFSMs to represent types described previously and calculate the probabilities for each observed data value. We now explain the corresponding PFSMs in detail.
\subsubsection*{Representing Types with PFSMs\label{sec:repr-struct-pfsm}}
Here, we show how a PFSM can be used to model a data type. We divide types into 2 groups: (i) primitive types consisting of integers, floats, Booleans and strings, and (ii) complex types such as IP addresses, email addresses, phone numbers, dates, postcodes, genders and URLs. The details regarding the implementation of the corresponding PFSMs can be found in Appendix \ref{sec:pfsm_data_type}.

\begin{figure}[!h]
\centering
\scalebox{0.9}{
\begin{tikzpicture}[>=stealth',shorten >=1pt,auto,node distance=3.5cm]
  \node[initial above,state] (q0)      {$q_0$};
  \node[initial above,state] (q1)  [right of=q0]     {$q_1$};
  \node[state, accepting]         (q2) [right of=q1]  {$q_2$};
  
  \path[->] (q0)  edge [bend left=30] node {$0.5(+)$} (q1) (q0) edge [bend right=30] node {$0.5(-)$} (q1)
(q1) edge [bend left=30] node {$0.1(0)$} (q2) (q1) edge node {\dots} (q2) (q2) (q1) edge [bend right=30] node {$0.1(9)$} (q2)
  (q2)  edge [in=65,out=115, loop] node {$p(0)$} (q2)  (q2)  edge [in=245,out=295, loop] node {$p(9)$} (q2) (q2) edge [in=335,out=25, loop]node {$\dots$} (q2) ;  
    
\end{tikzpicture}
}
\caption{Graphical representation of a PFSM with states $\theta = \{q_0, q_1, q_2\}$ and alphabet $\Sigma = \{\texttt{+}, \texttt{-}, 0, 1, \dots, 9\}$ where $p$ denotes $\frac{1-P_{stop}}{10}$.}
\label{fig:ptype_integer_pfsm_example}
\end{figure}
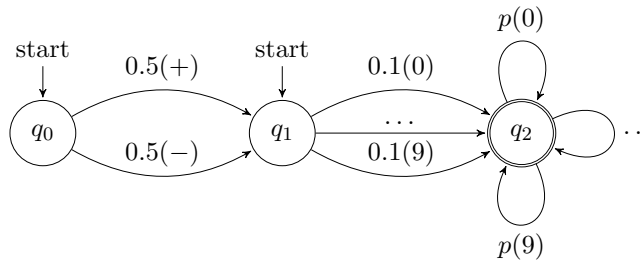

Consider integer numbers whose domain is $\{-\infty,\infty\}$. We can
represent the corresponding PFSM as in the diagram given in Figure
\ref{fig:ptype_integer_pfsm_example}. The machine has two initial
states, namely $q_0$ and $q_1$, and one final state $q_2$. Here, $q_0$
and $q_1$ respectively allow us to represent integer numbers with a
sign (plus or minus), or without any sign\footnote{The transitions
  from state $q_1$ allow the emission of a zero, which means that numbers like $-007$ can be emitted. If this is not desired then one can adjust the PFSM in Figure \ref{fig:ptype_integer_pfsm_example} to not emit a leading $0$.}.
The machine eventually transitions to the state $q_2$, which stops with a stopping probability $F(q_2) = P_{stop}$. Otherwise, it transitions to itself by emitting a digit with an equal probability, $(1-P_{stop})/10$. Similarly, we can develop a PFSM to represent each one of the other column types. 

The PFSM for missing values can be developed by using a predefined set of codes such as $\{\texttt{-1}, \texttt{-9}, \texttt{-99}, \texttt{NA}, \texttt{NULL}, \texttt{N/A}, \texttt{-}, \dots\}$. It assigns non-zero probabilities to each element in this set. Note that the corresponding PFSM already supports a wide range of codes, but it can be easily extended by the user to incorporate other known missing data encodings, leading to semi-automated operation.

In order to model anomalies, we adapt the idea of \emph{X-factor}
proposed by \mycitet{quinn2009factorial}. We define a
machine with the widest domain among all the PFSMs that supports all of the possible characters. This choice of PFSM lets the probabilistic model
become more robust against anomalies since it assigns probabilities to
data values that do not belong to any of the PFMSs representing the
data types. Note that as this PFSM covers a wider domain, it will assign
lower probabilities to known data types than the specific models.

Constructing PFSMs for complex types might require more human engineering than the other types. We reduce this need by building such PFSMs automatically from corresponding regular expressions. We first convert regular expressions to FSMs by using \mycitet{greenery} (see the function named to\_fsm()). We then build their probabilistic variants, where the parameters are assigned equal probabilities. Note that these parameters can be updated with the training.

\subsection{The Proposed Model \label{sec:model}}
We propose a new probabilistic mixture model with a noisy observation model, allowing us to detect missing and anomalous data entries. Our model first generates a column type from a set of possible regular data types. This is followed by a \say{deficiency} process that can potentially change the data type of each row. Consequently, each row might have a different type rather than the generated column type. The observation model then generates a data value for each entry according to its type. We now introduce our notation to represent this process.

We assume that a column of data $\textbf{x}=\{x_i\}_{i=1}^N$ has been read in,
where each $x_i$ denotes the characters in the $i^{th}$ row.\
We propose a generative model with a set of latent variables $t \in
\{1,2,...,K\}$ and $\textbf{z}=\{z_i\}_{i=1}^N$, where $t$ and $z_i$
respectively denote the data type of a column and its $i^{th}$
row. Here, $N$ is the number of rows in a data column and $K$ is the
number of possible data types for a column. We also use the additional
missing and anomaly types, denoted by $m$ and $a$ respectively, and
described above. Note that $z_i$ can be of type $m$ or $a$ alongside
a regular data type, i.e. $z_i \in \{1,2,...,K,m,a\}$. This noisy
observation model allows a type inference procedure robustified for
missing and anomalous data values.

Hence the model has the following generative process:
\begin{eqnarray*}
\text{column type $t$} &\sim& \mathcal{U}(1,K), \nonumber \\
\text{row type $z_i$}  &=&
\begin{cases} 
$t$ & \text{with probability $\pi_t^t$}\ ,\\
$m$ & \text{with probability $\pi_t^m$},\\
$a$ & \text{with probability $\pi_t^a$}\ ,\\
\end{cases} \nonumber\\
\text{row value $x_i$} &\sim& p(x_i | z_i),
\end{eqnarray*}
where $\boldsymbol{\Pi}$ and $p(x_i|z_i)$ are respectively the model parameter and
the observation model. $\mathcal{U}$ denotes a discrete Uniform
distribution. Here $\pi_t^t + \pi_t^m + \pi_t^a = 1$ for each column type $t$.
Since entries are often expected to be of a regular data type rather than missing or anomaly types, we favour regular types during inference by using lower coefficients for missing and anomaly types, i.e. $\pi_t^m < \pi_t^t$ and $\pi_t^a < \pi_t^t$. These weight parameters $\boldsymbol{\Pi}$ are assumed to be fixed and
known. Even though one can also learn the weights, this may not be vital as long as the coefficients of the regular types are larger than the others.

We use a uniform distribution for column types as in most cases we do
not have any prior information regarding the type of a column that
would allow us to favour a particular one. To represent the
conditional distribution $p(x_i | z_i)$, we have developed PFSMs as
described above. 

\subsection{Inference \label{sec:inference}}
Given a data column $\textbf{x}$, our initial goal is to infer the
column type $t$, which is cast to the problem of calculating the
posterior distribution of $t$ given $\textbf{x}$, denoted by
$p(t|\textbf{x})$. Then, we assume that each row can be of three
types: (i) the same as the column type, (ii) the missing type, and
(iii) the anomaly type. In order to identify missing or anomalous data entries, 
we calculate the posterior probabilities of each row type, namely $p(z_i|t,
\textbf{x})$. In this section, we now briefly discuss the corresponding calculations; detailed derivations are presented in Appendix \ref{sec:derivations_inference}.

Assuming that entries of a data column are conditionally independent given $t$, we can obtain the posterior distribution of column type $t$ as follows: 

\begin{equation}\label{eqn:posterior-column-type}
p(t=k|\textbf{x}) \propto p(t=k) \prod_{i=1}^N \Big( \pi_k^k p(x_i | z_i = k) + \pi_k^m p(x_i | z_i = m) + \pi_k^a p(x_i | z_i = a) \Big),
\end{equation}

which can be used to estimate the column type $t$, since the one with maximum posterior probability is the most likely data type corresponding to the column $\textbf{x}$. 

As per equation \ref{eqn:posterior-column-type}, the model estimates the column type by considering all the data rows, i.e. having missing data or anomalies does not confuse the type inference. Note that such entries would have similar likelihoods for each column type, which allows the model to choose the dominant data type for regular entries.

Following the inference of column type, we can also identify entries of $\textbf{x}$ which are more likely to be missing or anomalies rather than the inferred type. For this, we compare the posterior probabilities of each row type $z_i$ given $t=k$ and $x_i$, namely $p(z_i=j | t=k, x_i)$, which can be written as:
\begin{equation} \label{eq:posterior-row-type}
p(z_i=j | t=k, x_i) = \frac{\pi^{j}_{k} p(x_i | z_i=j)}{\sum_{\ell \in \{ k, m, a\}} \pi^{\ell}_{k} p(x_i | z_i=\ell)}.
\end{equation}

\subsubsection*{Complexity Analysis}
\label{sec:inference-complexity}
The computational bottleneck in the inference is the calculation of $p(\textbf{x} | t = k)$ for each type $k$, which is the calculation of the probability assigned for a data column $\textbf{x}$ by the $k^{th}$ PFSM. Note that this can be carried out by taking into account the counts of the unique data entries, for efficiency. Denoting the $u^{th}$ unique data value by $x_u$, we need to consider the complexity of calculating $p(x_u | t = k)$ which can be done via the PFSM Forward algorithm. Each iteration of the algorithm has the complexity of $O(M_k^2)$ where $M_k$ is the number of hidden states in the $k^{th}$ PFSM. As the number of iterations equals to the length of $x_u$ denoted by $L$, the overall complexity of the inference becomes $O(U K M^2 L)$, where $U$ is the number of unique data entries, $K$ is the number of types, and $M$ is the maximum number of hidden states in the PFSMs.

\subsection{Training of the Model \label{sec:training}}
The aim of the training is to tune the probabilities assigned by the
PFSMs so that column types are inferred accurately. Given a set of
columns and their annotated column types, the task is to find the
parameters of the PFSMs (i.e.\ the initial-state, the transition, and
the final-state probabilities) that allow the ``correct'' machine to give
higher probabilities to the observed entries. This is crucial as
multiple PFSMs can assign non-zero probabilities for certain
strings, e.g., $\texttt{1}$, $\texttt{True}$, etc.

We employ a discriminative training procedure on our generative model, as done in discriminative training of HMMs \mycitep{bahl1986maximum,jiang2010discriminative,brown1987acoustic,nadas1988model,williams1991mean}. This is shown to be generally superior to maximum likelihood estimations \mycitep{jiang2010discriminative}, since a discriminative criterion is more consistent with the task being optimized. Moreover, it allows us to update not only the parameters of the ``correct'' PFSM but also the parameters of the other PFSMs given a column of data and its type, which in turn helps the correct one to generate the highest probability.

We choose $\sum_{j=1}^M \log p(t^j|\mathbf{x}^j)$ as the objective function to maximize, where $\mathbf{x}^j$ and $t^j$ respectively denote the $j^{th}$ column of a given data matrix $X$ and its type, and $M$ is the number of columns. We then apply Conjugate Gradient algorithm to find the parameters that maximize this objective function (please see Appendix \ref{sec:derivations_training} for detailed derivations of the gradients).

We study different parameter settings for our model. We first explore tuning the parameters by hand to incorporate certain preferences over the types, e.g., Boolean over Integer for $\texttt{1}$. Then, we learn the parameters via the discriminative training described above where the parameters are initialized at the hand-crafted values. Note that due to the absence of explicit labels for the missing and anomaly types, these are not updated from the hand-crafted parameters. We have also employed the training by initializing the parameters uniformly. However, we do not report these results as they are not competitive with the others.

As the PFSMs are generative models it would be possible to train them
\emph{unsupervised}, to maximize $\sum_{i, j} \log p(x^j_i)$, where
$p(x^i_j)$ is defined as a mixture model over all types (including
missing and anomaly) for the $i$th row and $j$th column. The component
PFSMs could then be updated using the Expectation-Maximization (EM) algorithm. However, such
training would be unlikely to give as good classification performance
as supervised training.

\section{Related Work}
\label{sec:type_inference_related_work}
Numerous studies have attempted to tackle type inference, including wrangling tools and software libraries. Such approaches have made limited use of probabilistic methods, while the existing probabilistic approaches do not address the same type inference problem that we do. Below we discuss the capabilities of these existing methods in the presence of missing and anomalous data.

Type inference is commonly carried out by \emph{validation functions} and regular expressions. For example, several wrangling tools, including \mycitet{Trifacta2018} and its preceding versions \mycitep{Raman2001,Kandel2011,Guo2011} apply \emph{validation functions} to a sample of data to infer types, e.g., assign the one validated for more than half of the non-missing entries as the column type \mycitep{Kandel2011}. When multiple types satisfy this criterion, the more specific one is chosen as the column type, e.g., an integer is assumed to be more specific than a float. Trifacta supports a comprehensive set of data types, and provides an \emph{automatic discrepancy detector} to detect errors in data \mycitep{Raman2001}. However, in our experience, its performance on type inference can be limited on messy data sets.

\mycitet{fisher2005pads} and \mycitet{fisher2008dirt} have developed data description languages 
for processing ad hoc data (PADS) which enables generating a human-readable description of a dataset, based on data types inferred using regular expressions. However, their focus is on learning regular expressions to describe a
dataset, rather than classifying the data columns into known types. Unlike the PADS library which parses a data file itself, Test-Driven Data Analytics (TDDA, \citealt{tdda2018}) uses the Pandas CSV reader to read the data into a data frame. It then uses the Pandas dtypes attributes\footnote{obtained with the function pandas\_tdda\_type().}, to determine the data type of the columns. However, this leads to a poor type detection performance since the Pandas reader is not robust against missing data and anomalies, where only empty string, \texttt{NaN}, and \texttt{NULL} are treated as missing data.

\cite{fsharp-data-pldi2016} propose another use of regular expressions with F\#, where types, referred as \emph{shapes}, are inferred w.r.t.\ a set of \emph{preferred shape relations}. Such relations are used to
resolve ambiguous cases in which a data value fits multiple types. This procedure allows integrating inferred types into the
process of coding, which can be useful to interact with data. However,
it does not address the problems of missing and anomalous data
comprehensively, where only three encodings of missing data,
\texttt{NA}, \texttt{\#N\/A}, and \texttt{:}, are supported. This may lead to poor performance on type inference when missing data is encoded differently, or there are anomalies such as error messages.

A set of software packages in R and Python can also infer data types. messytables \mycitep{messytables2017} determines the most probable type for a column by weighting the number of successful conversions of its elements to each type. This can potentially help to cope with certain data errors; however, it might be difficult to find an effective configuration of the weights for a good performance. Moreover, it can not handle the \emph{disguised missing data} values, e.g., \texttt{-1} in a column of type integer, which can be misleading for the analysis. \mycitet{hypoparsr2017} propose a CSV parser named hypoparsr that treats type inference as a parsing step. It takes into account a wide range of missing data encodings; however, it does not address anomalies, leading to poor performance on type inference. readr \mycitep{readr2017} is an R package to read tabular data, such as CSV and TSV files. However, in contrast to hypoparsr, a limited set of values are considered as missing data, unless the user specifies otherwise. Furthermore, it employs a heuristic search procedure using a set of matching criteria for type inference. The details regarding the criteria are given in \mycitet{wickham2016r}. The search continues until one criterion is satisfied, which can be applied successfully in certain scenarios. Whenever such conditions do not hold, the column type is assigned to string.

\mycitet{pmlr-v70-valera17a} propose a model for discovery of
statistical types, but this is tackling a very different problem than
the one that we address.  Their method assumes that all of the entries
in a column contain numerical data values, i.e.\ it cannot handle data
types such as string, date, etc. Moreover, it can only handle clean
integer and float columns that do not have any missing data and
anomalies.  Given this they address the problem of making fine-grained
distinctions between different types of continuous variables
(real-valued data, positive real-valued data, and interval data) and
discrete variables (categorical data, ordinal data, and count data).
It would be possible to combine their method with ours to provide more fine-grained distinctions of real-valued and discrete-valued data once the issues of detecting missing and anomalous data have been resolved.

\mycitet{pmlr-v70-valera17a} is extended in \mycitet{Vergarietal19} where the proposed method can be used to impute missing data and
detect numerical anomalies. However, they also assume that each
entry of a data column contains a numerical value or an explicit
missing data indicator. Their method cannot be used when a data column contains non-numerical missing data encodings and anomalies. Moreover, the authors address a completely different missing data problem,
i.e. they tackle missing data imputation rather than missing data
detection. This means that the proposed method cannot detect whether an entry is
missing or not, but it can impute missing data once the location of
missing data is known. Lastly, their focus on anomaly detection is
also different than ours in the sense that they attempt to detect
numerical outliers, but do not detect string anomalies in a column.

\mycitet{limaye2010annotating} propose a log-linear model based method to
annotate a table in terms of the \emph{semantic} column type, cell
entity, and column relations, given an ontology of relevant
information. For example, given a table that contains information
about actors such as names and ages, the task is to find semantic
column types such as actor and age, the entities each row refers to,
and the relationship of actor-age between two columns. Even though
the scope of these annotations are wider than ours, missing data and
anomalies are not considered explicitly, and an appropriate
ontology is required.

\section{Experiments}
\label{sec:experiments}
In this section, we first in Sec.\ \ref{sec:exp-setup}
describe the datasets and evaluation metrics
used, and then in Sec.\ \ref{sec:quan-res}
compare \emph{ptype} with competitor methods on two tasks: (i) column type inference, and (ii) type/non-type inference. Lastly, we present a qualitative evaluation of our method on challenging cases in Sec.\ \ref{sec:qual-res}. The goal of our experiments is to evaluate (i) the robustness of the methods against missing data and anomalies for column type inference and (ii) the effectiveness of type/non-type inference. These are evaluated both quantitatively (Sec.\ \ref{sec:quan-res}) and qualitatively (Sec.\ \ref{sec:qual-res}). We release our implementation of the proposed method at \url{https://github.com/tahaceritli/ptype-dmkd}.

\subsection{Experimental Setup \label{sec:exp-setup}}
We have trained \emph{ptype} on 25, and tested on 43 data sets obtained
from various sources 
including UCI
ML\footnote{\url{https://archive.ics.uci.edu/ml/datasets.html}},
\url{data.gov.uk}, \url{ukdataservice.ac.uk}, and \url{data.gov}. 
The data types were annotated by hand for these sets. We also annotated each dataset in terms of missing data and anomalies, by using the available meta-data, and checking the unique values. 

On column type inference, we compare our method with F\# \mycitep{fsharp-data-pldi2016}, hypoparsr \mycitep{hypoparsrcode}, messytables \mycitep{messytables2017}, readr \mycitep{readr2017}, TDDA \mycitep{tdda2018} and \mycitet{Trifacta2018}. Note that some of the related works are not directly applicable for this task, and these are not included in these experiments. For example, we are not able to use \mycitet{Raman2001}, \mycitet{Kandel2011} and \mycitet{Guo2011} as they are not available anymore. However, we use their latest version Trifacta in our experiments. We also exclude the PADS library \mycitep{fisher2005pads,fisher2008dirt}, since it does not necessarily produce columns and their types. The methods proposed by \mycitet{pmlr-v70-valera17a} and \mycitet{Vergarietal19} are also not applicable for this task. First, they do not consider data types of Boolean, string, date. Secondly, they only address integer and float columns that do not contain any non-numerical missing data or anomalies, which are commonly found in real-world datasets. Note that \mycitet{Vergarietal19} do not address missing data detection but missing data imputation, and can only handle numerical outliers but not non-numerical outliers, whereas \mycitet{pmlr-v70-valera17a} do not address these questions. Lastly, we exclude the method presented by \mycitet{limaye2010annotating} as their goal is to infer semantic entity types rather than syntactic data types. 

On type/non-type inference, we evaluate the performance of our method in detecting missing data and anomalies. We label such entries as non-type, and classify each entry as either type or non-type. On this task, we compare our method with Trifacta only, as it is the leading competitor method on column type inference, and the others do not address this task comprehensively.

\subsubsection*{Data Sets}
We have conducted experiments on the data sets chosen according to two
criteria: (i) coverage of the data types, and (ii) data quality. 
In our experiments, we consider five common column types including Boolean, date, float, integer, and string. Table \ref{table:ptype_column_types} presents the distribution of the column
types found in our data sets. Any other columns not conforming to the 
supported data types are omitted from the evaluations.
Secondly, we have selected messy data sets in order
to evaluate the robustness against missing data and anomalies. The fraction of the non-type entries in the test datasets can be as large as 0.56, while the average fraction is 0.11. Note that available data sets, their descriptions and the corresponding annotations can be accessed via \url{https://goo.gl/v298ER}.

\begin{table}[h!]
\centering
\scalebox{0.8}{
\large
\begin{tabular}{lllllll}
\toprule 
\multicolumn{1}{c}{} &  \multicolumn{5}{c}{\bfseries Column Type} \\\cline{2-6}
 \addlinespace[1mm] 
  & Boolean &  Date & Float &  Integer &  String & Total\\
  \midrule
Training & 75 & 49 & 99 & 257 & 309 & 789 \\
Test & 43 & 40 & 53 & 240 & 234 & 610 \\
\bottomrule
\end{tabular}
}
\caption{Histogram of the column types observed in the training and the test data sets.}
\label{table:ptype_column_types}
\end{table}

\subsubsection*{Evaluation Metrics}
For column type inference, we first evaluate the overall accuracy of the methods on type inference by using the accuracy. However, this may not be informative enough due to the imbalanced data sets. Note that the task of type inference can be seen as a multi-class classification problem, where
each column is classified into one of the possible column types. In order to measure the performance separately for each type, we follow a \emph{one-vs-rest} approach. 
In such cases a common choice of metric is the Jaccard index $J$ (see e.g., \mycitet[sec 2.3]{hand2001}) defined as $\text{TP}/(\text{TP} + \text{FP} + \text{FN})$, where TP, FP, and FN respectively denote the number of True Positives, False Positives and False Negatives. (Note that \emph{one-vs-rest} is an asymmetric labelling, so True Negatives are not meaningful in this case.)

To measure the performance on type/non-type inference, we report Area
Under Curve (AUC) of Receiver Operating Characteristic (ROC) curves,
as well as the percentages of TPs, FPs, and FNs. Note that here we denote
non-type and type entries as Positive and Negative respectively.

\subsection{Quantitative Results \label{sec:quan-res}}
We present quantitative results on two tasks: (i) column type inference, and (ii) type/non-type detection. In column type inference, we evaluate the performance of the methods on detecting data types of columns, and investigate their scalability. Then, in type/non-type detection we evaluate their capability of detecting missing data and anomalies.

\subsubsection*{Column Type Inference}
We present the performance of the methods in Table
\ref{table:ptype_type_based_accuracies_training_new}, which indicates
that our method performs better than the others for all types, except for the date
type where it is slightly worse than Trifacta.   
In the table ptype denotes the discriminatively trained model, and ptype-hc the version with hand-crafted parameters. These improvements are
generally due to the robustness of our method against missing data and
anomalies.

Notice that the discriminative training improves the performance, specifically for Boolean and integer types. This shows that the corresponding confusions can be reduced by finding more optimal parameter values, which can be difficult otherwise. Note that the training has a slightly negative effect on float and string types, but it still performs better than the other methods. These changes can be explained by the fact that the cost function aims at decreasing the overall error over columns rather than considering individual performances.  

\begin{table}[h!]
\centering
\scalebox{0.75}{
\large
\begin{tabular}{llllllll}
\toprule 
\multicolumn{1}{c}{} &  \multicolumn{7}{c}{\bfseries Method} \\\cline{2-8} \addlinespace[1mm] 
  &   F\# & messytables &  readr & TDDA & Trifacta & ptype-hc & ptype \\
\midrule
Overall  & \multirow{2}{*}{0.73} &  \multirow{2}{*}{0.72} &  \multirow{2}{*}{0.69} & \multirow{2}{*}{0.61} &  \multirow{2}{*}{0.90} &  \multirow{2}{*}{0.92} &  \multirow{2}{*}{\bfseries  0.93} \\
Accuracy &&&&&& \\
\hline 
\addlinespace[1mm] 
Boolean & 0.55 &  0.56  &  0.00  &  0.00 &  0.49  & 0.75 & \bfseries 0.83 \\
Date    & 0.35 &  0.17  &  0.10 &  0.00  & \bfseries 0.68 &  0.67 & 0.67 \\
Float   & 0.60 &  0.57  & 0.59 & 0.42 & 0.87 & \bfseries 0.93 & 0.91 \\
Integer & 0.55 &  0.55  & 0.57 & 0.46 & \bfseries 0.88 & 0.85 & \bfseries 0.88 \\
String  & 0.61 &  0.61 & 0.58 & 0.51 & 0.83  &  \bfseries 0.90 & 0.89 \\
\bottomrule
\end{tabular}
}
\caption{Performance of the methods using the Jaccard index and overall accuracy, for the types Boolean, Date, Float, Integer and String.}
\label{table:ptype_type_based_accuracies_training_new}
\end{table}

\begin{figure}[ht!]
	\centering	
	\begin{subfigure}[b]{0.3\textwidth}
    \centering
		\includegraphics[width=\textwidth]{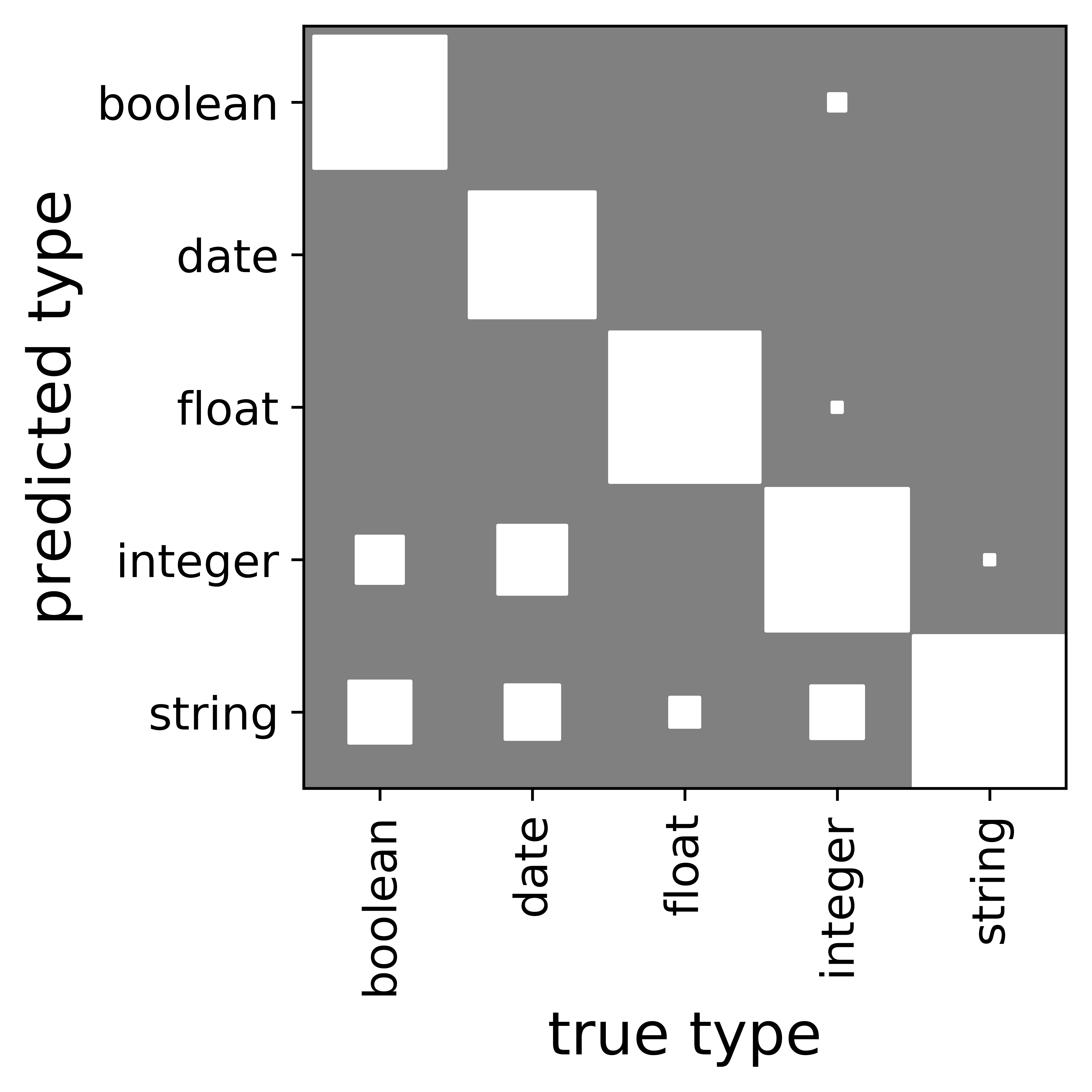}
		\caption{ptype}
	\end{subfigure}%
	\begin{subfigure}[b]{0.3\textwidth}
    \centering
		\includegraphics[width=\textwidth]{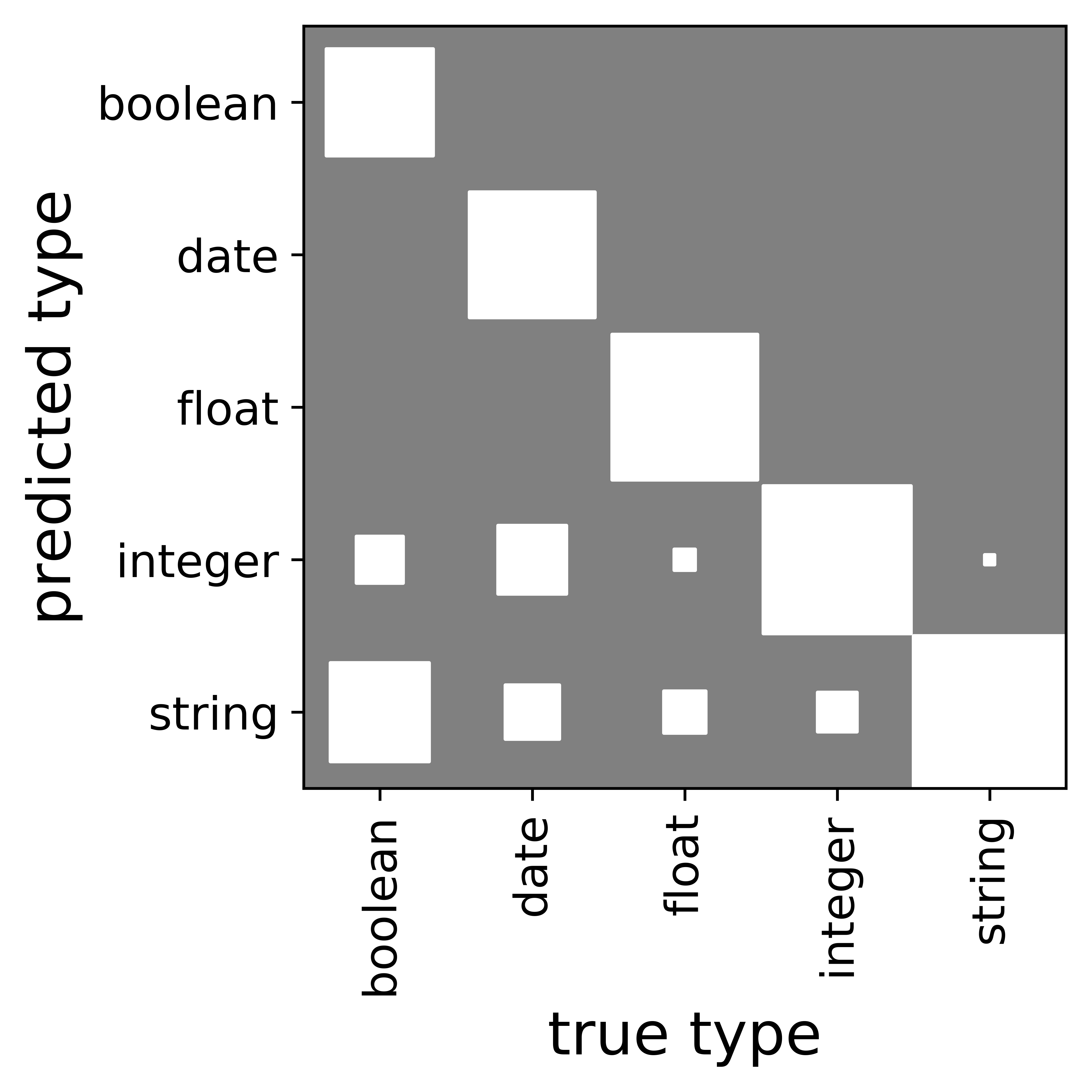}
		\caption{Trifacta}
	\end{subfigure}    
	\caption{Normalized confusion matrices for (a) ptype and (b) Trifacta plotted
as Hinton diagrams, where the area of a square is proportional to the
magnitude of the entry.}
	\label{fig:ptype-trifacta-new}
\end{figure}

Figure \ref{fig:ptype-trifacta-new} shows normalized confusion matrices for ptype and Trifacta, normalized so that a column sums to $1$. This shows that they both tend to infer other column types as string even if they are not. However, ptype has few of such confusions, especially when the true type is Boolean or float.

It is noticeable from Table \ref{table:ptype_type_based_accuracies_training_new} that dates are difficult to infer accurately. Detailed inspection shows that this is due to non-standard formats used to denote dates. We could improve the PFSM for dates in \emph{ptype} to include such formats, but we have not done so, to avoid optimizing on test datasets. The performance of Trifacta on dates can be explained by the engineering power behind the tool, and indicates its capability to represent non-standard formats using validation functions.

To determine whether the column type predictions of ptype and Trifacta
are significantly different, we apply the McNemar's test
(see e.g.,  \citealt*{dietterich1998approximate}), which assumes that the two
methods should have the same error rate under the null hypothesis. 
We compute the test statistic
$(\abs{n_{01}-n_{10}}-1)^2)/(n_{01}+n_{10})$, where $n_{01}$ and
$n_{10}$  denote 
the number of test columns misclassified by only Trifacta, and by only
ptype respectively. In our case, $n_{01}$ and $n_{10}$ are respectively equal to 19 and 6,
which results in a statistic of 5.76. 
If the null hypothesis is correct, then the probability that this
statistic is greater than 3.84 is less
than 0.05 \mycitep{dietterich1998approximate}. Thus this result
provides evidence to reject the null hypothesis, and confirms that
the methods are statistically significantly different to each other.

The large performance gap for Booleans suggests that our method handles confusions with integers and strings better. Analysis shows that such confusions occur respectively in the presence of 
$\{\texttt{0},\texttt{1}\}$, and $\{\texttt{Yes}, \texttt{No}\}$\footnote{We assume that a data column where the entries are valued as Yes or No is more likely to be a Boolean column than a string. We have also confirmed these cases with the corresponding metadata whenever available, and have carefully annotated our datasets in terms of data types.}. We further note that F\#, messytables, and readr perform similarly, especially on floats, integers, and strings; which is most likely explained by the fact that they employ similar heuristics. 

Since hypoparsr names column types differently, except for the date type, we need to rename the annotations, and evaluate the methods again in order to compare them with hypoparsr. It refers to Boolean and string respectively as logical and text. Moreover, integer and float are grouped into a single type called numeric. The resulting evaluations, which are reported in Table \ref{table:evaluations_hypoparsr} in Appendix \ref{sec:additional_exp_results}, shows a similar trend to as before in that our method performs better. However, we see some variations, which result from the fact that we now evaluate on a smaller number of data sets since hypoparsr, which is said to be designed for small sized files, was able to parse only  33 out of the 43 test data sets\footnote{This was using the default settings, but Till D\"{o}hmen (pers.\ comm.) advised against changing the limit.}. This left us 358 columns including 29 date, 21 logical, 159 numeric, and 149 text. Lastly, we observe that hypoparsr results in a higher number of confusions by inferring the type as integer whereas the true type is text. Such cases mostly occur when the data values consist of a combination of numeric and text, e.g., ID and address columns.

\subsubsection*{Scalability}
We describe the complexity of the inference in our model in Section \ref{sec:inference-complexity}. Here, we demonstrate its scaling by measuring the time it takes to infer the column types for each test dataset. 

\begin{figure}[ht!]
\centering	
\includegraphics[width=.6\textwidth]{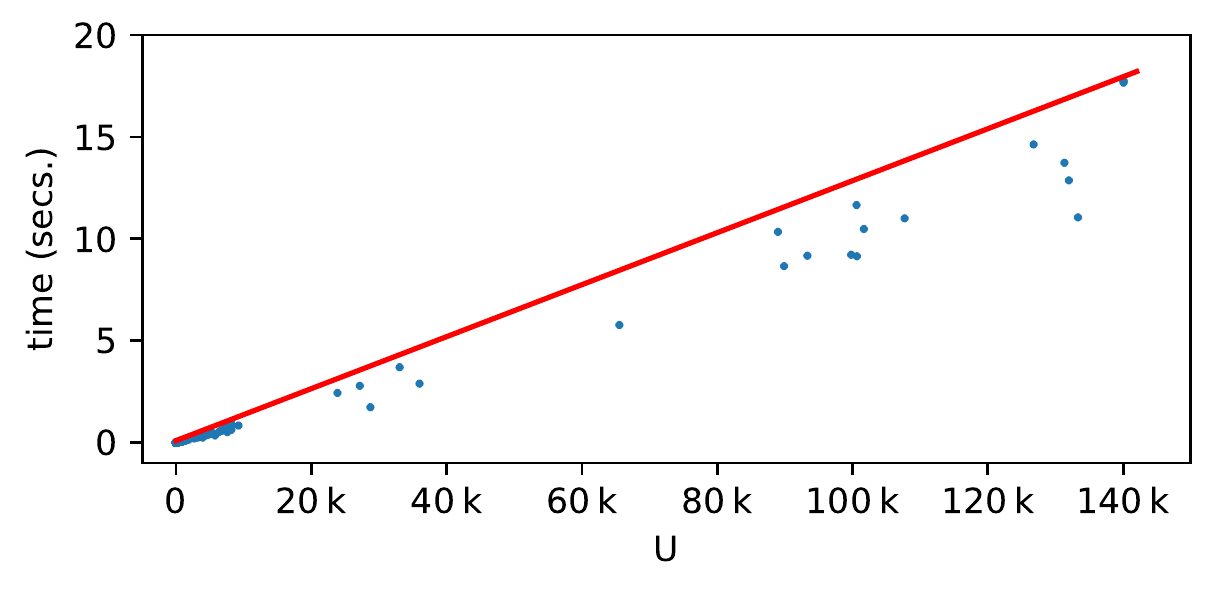}		
\caption{The time in seconds taken to infer a column type with ptype, as a function of the number of unique values $U$ in the column. Also shown is the line $c_0 + c_1 U$, where $c_0$ is a small constant.}
\label{fig:ptype_time_takes}
\end{figure}

Recall that the complexity is $O(U K M^2 L)$, where $U$ is the number of unique data entries, $K$ is the number of data types, $M$ is the maximum number of hidden states in the PFSMs, and $L$ is the maximum length of data values. Notice that the complexity depends on data through $U$ and $L$, and does not necessarily increase with the number of rows. In fact, it grows linearly with the number of unique values assuming $L$ is constant. As shown in Fig.\ \ref{fig:ptype_time_takes}, the runtime for ptype is upper bounded by a line
$c_0 + c_1 U$, where $c_0$ is a small constant. The runtime thus
scales linearly with the number of unique values $U$, handling
around 10K unique values per second. The variations can be
explained by changes in L.

We also report the size of the datasets and the times the methods take in Appendix \ref{sec:scalability_appendix}. We have observed similar performance with messytables, whereas readr and TDDA seem much faster even though they do not only predict the data types but also parse a given dataset. On the other hand, hypoparsr takes much longer compared to the others. Lastly, we measure the processing times for Trifacta via command line. We have observed that Trifacta takes less time than hypoparsr, but it often takes longer compared to the other methods.

\subsubsection*{Type/Non-type Inference}
Trifacta labels each entry in a data table either as type or non-type, whereas our model presents a probability distribution over the two labels. One could apply a threshold on these probabilities in order to assign a label to each entry. Here, we demonstrate how the methods behave under different thresholds. We aggregate the entries of each dataset over its columns, and compute the ROC curve for each method.

\begin{figure}[ht!]
  \centering
    \includegraphics[width=.7\textwidth]{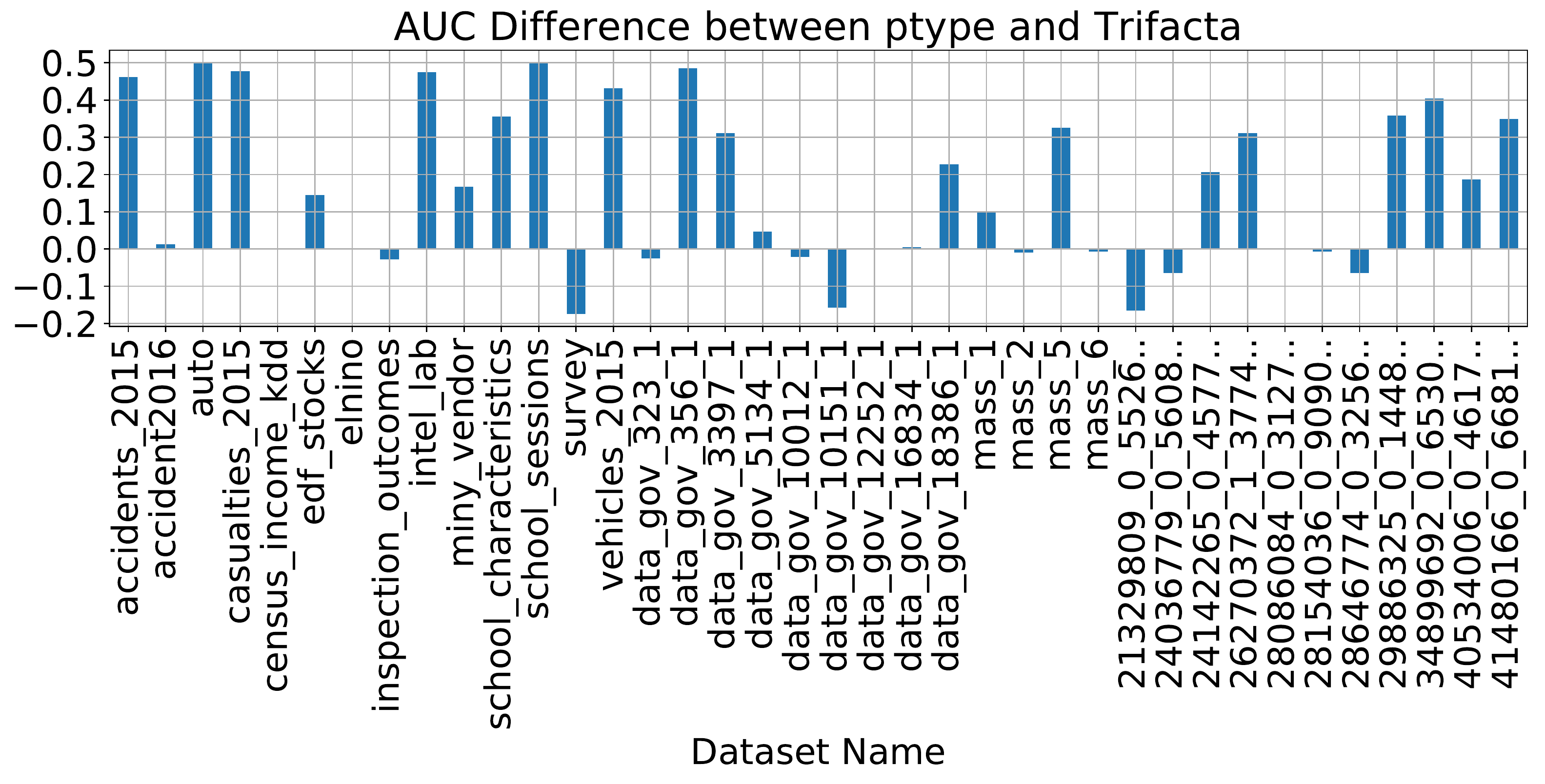}
  \caption{AUC(ptype) - AUC(Trifacta) plotted for each test dataset.}
\label{fig:aucs_dataset}
\end{figure}

Figure \ref{fig:aucs_dataset} presents the difference AUC(ptype) - AUC(Trifacta)
per dataset. Note that we exclude five datasets as the corresponding AUCs are undefined due to the definition of True Positive Rate ($\frac{TP}{TP+FN}$). This becomes undefined since the denominator becomes zero when both TP and FN are equal to zero, which occurs naturally when a dataset does not contain any missing data and anomalies. The average of AUCs of the remaining datasets are respectively 0.77 and 0.93 for Trifacta and ptype. To compare these two sets of AUCs, we apply a paired t-test, which results in the t-statistic of 4.59 and p-value of 0.00005. These results reject the null hypothesis that the means are equal, and confirm that they are significantly different.

Lastly, we compare Trifacta and ptype in terms of percentages of TPs, FPs, and FNs which are presented in Table \ref{table:type_nontype_evaluations_numbers}, where the labels for ptype are generated by applying a threshold of 0.5 on the posterior distributions. Note that here we aggregate the predictions over the datasets. As per the table, ptype results in a higher number of FPs than Trifacta, but Trifacta produces a higher number of FNs and a lower number of TPs than ptype. Note that here we denote non-type and type entries as Positive and Negative respectively.\ This indicates that ptype is more likely to identify non-types than Trifacta, but it can also label type entries as non-types more often. However, the overall performance of ptype is better than Trifacta, as we have also observed in the AUCs.

\begin{table}[h!]
\centering
\begin{tabular}{l|lll}
\toprule 
Method & FPs & FNs & TPs \\
  \midrule
Trifacta & 0.67 & 3.96 & 1.57 \\
ptype & 1.13 & 0.20 & 5.34\\ 
\bottomrule
\end{tabular}
\caption{The percentages of FPs, FNs, and TPs for Trifacta and ptype on type/non-type detection.}
\label{table:type_nontype_evaluations_numbers}
\end{table}

We now present two examples to give insight into Table \ref{table:type_nontype_evaluations_numbers}. Consider the ``Substantial\_growth\_of\_\\knowledge\_skills'' column of the mass\_6 dataset which consists of floating-point numbers. However, most of the 3148 entries are non-types, i.e.\  empty entries, \texttt{-}, \texttt{N/A}, and \texttt{NA} which occur 780, 470, 1063, and 424 times respectively. Such non-type entries are labeled correctly by ptype, whereas Trifacta can only classify the empty entries correctly as non-type. The remainder of the non-type entries are considered to be valid type entries, since they conform with the column type which is inferred as string by Trifacta. Note that this confusion is high likely due to the low number of floats in the column. Here, Trifacta results in a high percentage of FNs, and a lower percentage of TPs than ptype.

The second example illustrates the extent of cases in which ptype results in FPs. For example, the ``CAUSE\_NAME'' column in the data\_gov\_323\_1 dataset consists of data values such as \texttt{Cancer}, \texttt{Stroke}, and \texttt{Suicide} etc. Here, ptype and Trifacta infer the column type as string, and label such entries correctly as type entries. However, \texttt{Alzheimer's disease} and \texttt{Parkinson's disease} are misclassified as non-types by ptype (1,860 FPs out of 13,260 entries) as our string model does not support the apostrophe. To handle this, we could include \texttt{'} in the corresponding alphabet, but we also find it helpful to detect ``true'' non-type entries having that character. We believe that such cases should be left to users with domain knowledge as they can easily extend the alphabet of the string type.

\subsection{Qualitative Results \label{sec:qual-res}}
We now give some examples of predicting missing and anomalous data.

\begin{figure}
  \centering
    \includegraphics[width=.5\textwidth]{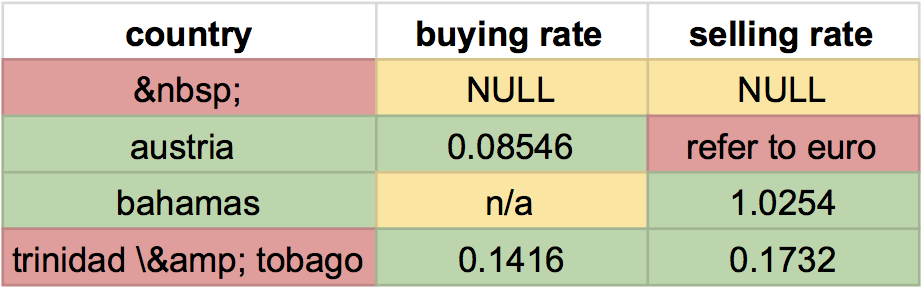}
  \caption{Annotating a subset of a T2D data set using \emph{ptype} as \textcolor{green}{\textbf{normal}}, \textcolor{yellow}{\textbf{missing}}, and \textcolor{red}{\textbf{anomalous}} entries, denoted by green, yellow, and red, respectively.}
\label{fig:ptype_example_anomaly}
\end{figure}

\underline{Missing Data:} We support an extensive range of values that are used to denote missing data. Note that multiple such encodings can be detected at the same time. Consider a T2Dv2 dataset\footnote{34899692\_0\_6530393048033763438.csv} where missing entries are denoted by encodings such as \texttt{NULL} and \texttt{n/a}. Our method can successfully annotate such entries as shown in Figure \ref{fig:ptype_example_anomaly}. 

Next, we show how our model approaches to unseen missing data encodings which are not explicitly considered as missing data in our model, but can be handled with the anomaly type. For example, ``Time of Incident'' column of Reported Taser 2015 data set is expected to contain date values. However, some entries have the value \texttt{Unknown}. Thanks to the PFSM for anomaly type, our model detects such encodings as anomalies as long as the \say{correct} column type is not string, resulting in a better performance of type inference.

Another advantage of \emph{ptype} is that it can detect missing data
even if their types fit the column type. Consider an integer-typed
column Stake in the Rodents data set where \texttt{-99} is used to denote missing entries. Our method flags those entries as missing instead of treating them as regular integers since the missing type accepts \texttt{-99}. Similarly, we are able to detect string-valued missing data in string-typed columns. When the column type however is not string, our model may result in a number of false alarms by flagging normal entries as missing data. For example, integer values of \texttt{-1}, \texttt{-99}, etc. can also represent normal data instances. We will investigate this using methods as in \cite{qahtan2018fahes}, and develop missing
data models specific to each data type, in order to improve this issue.

\underline{Anomalies:} As mentioned earlier, we can also use \emph{ptype} to detect anomalies. Our model flags such entries automatically since the anomaly model covers a wide range of values including those that are not supported by the column type. 

Figure \ref{fig:ptype_example_anomaly} shows the capability of
detecting anomalies when the column type is string. As we do not have
the character \texttt{\&} in the alphabet of the PFSM for strings, the
anomaly machine allows us to detect the anomalies in the ``country''
column. Similarly, the characters ``\texttt{refer to euro}'' are not
supported by the PFSM for integers, letting us detect the corresponding
entry as anomalous. Moreover, we can separately detect the anomalous and missing data as in the ``selling rate'' column. 

Interestingly, we notice that the question mark character \texttt{?} is used to express the doubt about the collected data in the HES data set, where a data entry contains the value of \texttt{6?}. We can also see that missing data encodings not incorporated to our missing data model such as \texttt{NOT AVAILABLE}, \texttt{??} (double question marks), and \texttt{--}, are detected as anomalies. 

We now discuss some other aspects of our model, such as ambiguities that can occur, and how they can be handled; and failure cases which can be difficult to avoid.

\subsubsection*{The Limitations of Our Work}
\underline{Ambiguous Cases:}
In certain cases the posterior probability distribution over types is
not heavily weighted on a particular type. For example, consider a column of
data that contains values of \texttt{NULL} and \texttt{1}. This could
fit multiple PFSMs as \texttt{1} can either be an integer, a float, a
string, or a Boolean value. However, we have assumed that \texttt{0} and
\texttt{1} are more likely to be indicators of Booleans and have thus
tuned the parameters of the corresponding machine such that the
probabilities associated with \texttt{0} and \texttt{1} are slightly
higher than the ones assigned by the other machines. This leads to a
posterior probability distribution with values of 0.29, 0.26, and 0.44
respectively for integers, floats, and Booleans.  One can also exploit the
probabilistic nature of our model to treat such ambiguous cases
differently. Instead of directly taking the type with the highest
posterior probability as the column type, one can detect such
ambiguities, and then exploit user feedback to improve the decision.

A uniformly distributed posterior distribution over types is observed when
all of the entries of a column are assigned zero probabilities by the
PFSMs of the regular data types. This is not surprising as we only
support a limited set of characters in the regular machines, i.e. the
widest alphabet among the regular data types, which is of the string
type, consists of the letters, digits, and a set of punctuations. For
example, \say{per capitagdp (us\$)[51]} column of a T2D data
set\footnote{24036779\_0\_5608105867560183058.csv} has values such as
\texttt{\$2949.57}. Similarly, the 23rd column of the fuel data set
contains values such as \say{\texttt{( 3.0L) }}. Note that the anomaly type still assigns positive probabilities to these entries, as its alphabet includes all the possible characters. However, when its weight $\pi^a_t$ is the same regardless of the type, the corresponding posterior probabilities become equal. One can handle such cases by learning different weight parameters for the types. However, here, to represent such
unsupported/unknown types, we have developed \emph{X-type}, an
anomaly-like PFSM which is used as a regular column type. This can be
further exploited to create new data types, which can be automated via
a probabilistic training procedure.

\underline{Failure Cases:}
We now present two cases for which ptype-hc fails to infer the column
types correctly. For example, consider a column (\say{BsmtHalfBath} of
Housing Price data set) which denotes the number of half bathrooms in
the basement of houses, consisting of the values in $\{\texttt{0},
\texttt{1}, \texttt{2}\}$. In this case, ptype-hc puts higher posterior probability on the Boolean type whereas the actual type is integer. This may not be
surprising, considering the fact that \texttt{0} and \texttt{1} have
higher probabilities of being a Boolean, and \texttt{2}, occurring
only twice out of 1460 entires, is treated as an anomaly. However, ptype is able to correct this failure thanks to the discriminative training. Note that
the competitor methods fail in this case.

After the evaluations, we have discovered a set of cases we have not considered in the beginning. For example, several Boolean columns of the Census Income KDD data set have leading whitespace as in
``\texttt{ No}'', ``\texttt{ Yes}''. Our model infers the types of these Boolean columns as string since such whitespace is not considered in the Boolean type. In order to avoid optimizing the model on the test sets, we have not addressed such cases. However, they can easily be handled by updating the corresponding PFSMs to include the whitespace. Note that the other methods also detect the column types as string in these cases. 

There are cases of non-standard dates we do not currently
handle. For example, dates are sometimes divided into multiple columns as day, month, and
year. Our model detects day and month columns separately as integers. One could
develop a model that checks for this pattern, making use of 
constraints on valid day, month and year values.

\section{Summary}
\label{sec:summary}
We have presented ptype, a probabilistic model for column type
inference that can robustly detect the type of each column in a given
data table, and label non-type entries in each column. The proposed
model is built on PFSMs to represent regular data types
(e.g., integers, strings, dates, Booleans, etc.), missing and anomaly
types. An advantage of PFSMs over regular expressions is their ability
to generate weighted posterior predictions even when a column of
data is consistent with more than one type model. We have also
presented a discriminative training procedure which helps to improve
column type inference. Our experiments have demonstrated that we
generally achieve better results than competitor methods on messy data
sets.

Future work includes extending the supported data types, such as
categorical data, etc.; developing subtypes, e.g., for 
Booleans expecting either \texttt{True} and \texttt{False}, or
\texttt{yes} and \texttt{no}; and improving anomaly detection for
string-typed data by addressing semantic and syntactic errors.

\section*{Acknowledgements}
TC is supported by a PhD studentship from the Alan Turing Institute, under the EPSRC grant TU/C/000018. CW and JG would like to acknowledge the funding provided by the UK Government's Defence \& Security Programme in support of the Alan
Turing Institute. The work of CW is supported in part by EPSRC grant EP/N510129/1 to the Alan Turing Institute. We thank the anonymous reviewers for their comments that have helped
improve the paper.

\bibliographystyle{apalike} 
\bibliography{references}    

\appendix

\section*{Appendices}
In this Appendices, we discuss the implementation of the Probabilistic Finite State Machines (PFSMs) (Appendix \ref{sec:pfsm_data_type}), describe the data sets used (Appendix \ref{sec:pfsm_datasets}) and present the derivations for training and inference in our model (Appendix \ref{sec:derivations_training} and \ref{sec:derivations_inference}, respectively). Moreover, we present additional experimental results (Appendix \ref{sec:additional_exp_results}) and report scalability of the methods (Appendix \ref{sec:scalability_appendix}).

\section{PFSMs for Data Types}
\label{sec:pfsm_data_type}
In this work, we use five regular data types including integers, strings, floats, Booleans, and dates; and two noisy data types, namely missing and anomaly. 

\subsection{Integers}
Please see Sec.\ \ref{sec:repr-struct-pfsm} for a detailed discussion of the PFSM used to represent integers.

\subsection{Floats}
A floating-point number often consists of digits and a full stop character, which is followed by another set of digits. However, they can also be written without any fractional component, i.e.\ as integer numbers. We also support the representations of floating-point numbers with \texttt{e} or \texttt{E}. Lastly, we support the use of comma for the thousands separator in
floating-point numbers, such as \emph{1,233.15}, \emph{1,389,233.15}, etc.

\subsection{Strings}
The string PFSM is constructed with one initial state and one final state. Through each transition, either a digit, an alpha character, or a punctuation character is emitted. The punctuation characters considered here are \texttt{.}, \texttt{,}, \texttt{-}, \texttt{\_}, \texttt{\%}, \texttt{:}, and \texttt{;}, which are commonly found in real-world data sets to represent columns with the string type. 

\subsection{Booleans}
Our machine supports the following values by assigning them non-zero probabilities: \texttt{Yes}, \texttt{No}, \texttt{True}, \texttt{False}, \texttt{1}, \texttt{0}, \texttt{-1} and their variants \texttt{yes}, \texttt{Y}, \texttt{y}, \texttt{no}, \texttt{true}, \texttt{false}.

\subsection{Dates}
We categorize date formats into two groups, which are detailed below:

\subsubsection*{ISO-8601}
We support values in \emph{YYYY-MM-DDTHH:MM::SS}, where T is the time designator to indicate the start of the representation of the time of day component. We also support other ISO-8601 formats such as \emph{YYYYMMDD}, \emph{YYYY-MM-DD}, \emph{HH:MM}, and \emph{HH:MM:SS}. 

\subsubsection*{Nonstandard Formats}
We treat years in \emph{YYYY} format as date type. To distinguish years and integers, we restrict this to the range of [1000-2999]. On the other hand, we do not explicitly constrain the month (\emph{MM}) and day columns (\emph{DD}) to valid ranges, and but treat them as integers. We support ranges of years with the formats of \emph{YYYY}-\emph{YYYY}, \emph{YYYY}\ \emph{YYYY}, \emph{YYYY} - \emph{YYYY}, \emph{YYYY} -\emph{YYYY}, and \emph{YYYY}- \emph{YYYY}. Lastly, We support dates written as \emph{MM-DD-YYYY HH:MM:SS AM/PM}, and months, e.g., January, February, etc.

\subsection{Missing}
The machine for missing data assigns non-zero probabilities to the elements of this set, including \texttt{Null}, \texttt{NA} and their variants such as \texttt{NULL}, \texttt{null}, \say{\texttt{NA }}, \texttt{ NA}, \say{\texttt{N A}}, \texttt{N/A}, \say{\texttt{N/ A}}, \say{\texttt{N /A}}, \texttt{N/A}, \texttt{\#NA}, \texttt{\#N/A}, \texttt{na}, \say{\texttt{ na}}, \say{\texttt{na }}, \say{\texttt{n a}}, \texttt{n/a}, \texttt{N/O}, \texttt{NAN}, \texttt{NaN}, \texttt{nan}, \texttt{-NaN}, and \texttt{-nan}; special characters such as \texttt{-}, \texttt{!}, \texttt{?}, \texttt{*}, and \texttt{.}; integers such as \texttt{0}, \texttt{-1}, \texttt{-9}, \texttt{-99}, \texttt{-999}, \texttt{-9999}, and \texttt{-99999}; and characters denoting empty cells such as \say{\texttt{}} and \say{\texttt{ }}. 

\subsection{Anomaly}
We use all of the Unicode characters in this machine's alphabet, including the accented characters. Note that the number of elements in this set is 1,114,112.

\section{Data Sets}
\label{sec:pfsm_datasets}
We share the available data sets and the corresponding annotations at \url{https://goo.gl/v298ER}. Here, we briefly describe these data sets, and provide a list in Table \ref{table:datasets} which denotes their sources, and whether they are used in the training or testing phase.

\begin{itemize}
\item Accident 2016: information on accidents casualties across Calderdale, including location, number of people and vehicles involved, road surface, weather conditions and severity of any casualties.
\item Accidents 2015: a file from Road Safety data about the circumstances of personal injury road accidents in GB from 1979.
\item Adult: a data set extracted from the U.S.\ Census Bureau database to predict whether income exceeds \$50K/yr.
\item Auto: a data set consisting of various characteristics of a car, its assigned insurance risk rating, and its normalized losses in use.
\item Broadband: annual survey of consumer broadband speeds in the UK.
\item Billboard: a data set on weekly Hot 100 singles, where each row represents a song and the corresponding position on that week's chart.
\item Boston Housing: a data which contains census tracts of Boston from the 1970 census.
\item BRFSS: a subset of the 2009 survey from BRFSS, an ongoing data collection program designed to measure behavioral risk factors for the adult population.
\item Canberra Observations: weather and climate data of Canberra (Australia) in 2013.
\item Casualties 2015: a file from Road Safety data about the consequential casualties.
\item Census Income KDD: a data set that contains weighted census data extracted from the 1994 and 1995 current population surveys conducted by the U.S. Census Bureau. 
\item CleanEHR (Critical Care Health Informatics Collaborative): anonymised medical records \footnote{a subset is available at \url{https://github.com/ropensci/cleanEHR/blob/master/data/sample\_ccd.RData}}.
\item Cylinder Bands: a data set used in decision tree induction for mitigating process delays known as \say{cylinder bands} in rotogravure printing.
\item data.gov: 9 CSV files obtained from data.gov, presenting information such as the age-adjusted death rates in the U.S., Average Daily Traffic counts, Web traffic statistics, the current mobile licensed food vendors statistics in the City of Hartford, a history of all exhibitions held at San Francisco International Airport by SFO Museum, etc.
\item EDF Stocks: EDF stocks prices from 23/01/2017 to 10/02/2017.
\item El Ni\~{n}o: a data set containing oceanographic and surface meteorological readings taken from a series of buoys positioned throughout the equatorial Pacific.
\item FACA Member List 2015: data on Federal Advisory Committee Act (FACA) Committee Member Lists.
\item French Fries: a data set collected from a sensory experiment conducted at Iowa State University in 2004 to investigate the effect of using three different fryer oils on the taste of the fries.
\item Fuel: Fuel Economy Guide data bases for 1985-1993 model.
\item Geoplaces2: information about restaurants (from UCI ML Restaurant \& consumer data).
\item HES (Household Electricity Survey): time series measurements of the electricity use of domestic appliances (to gain access to the data, please register at \url{https://tinyurl.com/ybbqu3n3}).
\item Housing Price: a data set containing 79 explanatory variables that describe (almost) every aspect of residential homes in Ames, Iowa.
\item Inspection Outcomes: local authority children’s homes in England - inspection and outcomes as at 30 September 2016.
\item Intel Lab: a data set collected from sensors deployed in the Intel Berkeley Research lab, measuring timestamped topology information, along with humidity, temperature, light and voltage.
\item mass.gov: 4 CSV files obtained from mass.gov, which is the official website of the Commonwealth of Massachusetts.
\item MINY Vendors: information on \say{made in New York} Vendors.
\item National Characteristics: information on the overall, authorised, unauthorised and persistent absence rates by pupil characteristics.
\item One Plus Sessions: information on the number of enrollments with one or more session of absence, including by reason for absence.
\item Pedestrian: a count data set collected in 2016, that denotes the number of pedestrians passing within an hour.
\item PHM Collection:  information on the collection of Powerhouse Museum Sydney, including textual descriptions, physical, temporal, and spatial data as well as, where possible, thumbnail images.
\item Processed Cleveland: a data set concerning heart disease diagnosis, collected at Cleveland Clinic Foundation (from the UCI ML Heart Disease Data Set).
\item Sandy Related: Hurricane Sandy-related NYC 311 calls.
\item Reported Taser 2015: a hand-compiled raw data set based on forms filled out by officers after a stun gun was used in an incident, provided by CCSU's Institute for Municipal and Regional Policy.
\item Rodents: the information collected on rodents during a survey.
\item Survey: a data set from a 2014 survey that measures attitudes towards mental health and frequency of mental health disorders in the tech workplace. 
\item TAO: a real-time data collected by the Tropical Atmosphere Ocean (TAO) project from moored ocean buoys for improved detection, understanding and prediction of El Ni\~{n}o and La Ni\~{n}a.
\item Tb: a tuberculosis dataset collected by the World Health Organisation which records the counts of confirmed tuberculosis cases by \say{country}, \say{year}, and demographic group.
\item Tundra Traits: measurements of the physical characteristics of shrubs in the arctic tundra.
\item T2Dv2 Gold Standard: a set of data Web tables to evaluate matching systems on the task of matching Web tables to the DBpedia knowledge base.
\item User Profile: information about consumers (from UCI ML Restaurant \& consumer data).
\item Vehicles 2015: a file from Road Safety data about the types of vehicles involved in the accidents.
\item 83492acc-1aa2-4e80-ad05-28741e06e530: a hypoparsr data set which contains information on expenses.
\end{itemize}
Note that the data sets from mass.gov and data.gov are obtained from Abdulhakim A.\ Qahtan and also used in [16].
\begin{table}[ht!]
\center
\scalebox{0.7}{
\large
\begin{tabular}{lllll}
\toprule 
\bfseries name & \bfseries source & \bfseries{training/test} & \bfseries{\# columns} & \bfseries{\# rows} \\
\midrule
Accidents 2015 &  data.gov.uk & test & 32 & 140,056\\
Accident 2016  &  data.gov.uk & test  & 18 & 555\\
Adult & UCI ML & training & 15 & 32,561\\
Auto & UCI ML & test  &  26 & 205\\
Broadband & data.gov.uk & training  & 55 & 2,732\\
Billboard & github.com & training  & 72 & 317\\
Boston Housing & Kaggle & training & 15 & 333\\
Brfss & github.com  & training & 34 & 245 \\
Canberra Observations & others & training & 13 & 19,918 \\
Casualties 2015 & data.gov.uk & test & 16 & 186,189\\
Census Income KDD & UCI ML & test & 42 & 199,523\\
CleanEHR & others & training &  62 & 1,979 \\
Cylinder Bands & UCI ML & training & 40 & 540 \\
EDF Stocks & github.com & test & 7 & 5,425\\
Elnino & UCI ML & test & 9 & 782\\
FACAMemberList2015 & github.com & training & 21 & 72,220\\
French Fries & github.com & training & 10 & 696\\
Fuel & github.com & training & 35 & 941\\
Geoplaces2 & UCI ML & training & 20 & 130\\
HES & ukdataservice.ac.uk & training & 65 & 4,600\\
Housing Price & Kaggle & training & 81 & 1,460\\
Intel Lab & others & test  & 8 & 1,048,576\\
Inspection Outcomes & others & test & 22 & 1,477\\
MINY Vendor & data.gov & test & 18 & 897\\
Pedestrian                       & others & training & 9 & 37,700\\
Phm              	        	 & others & training & 16 & 75,814\\
Processed Cleveland 		     & UCI ML & training & 14 & 303\\
Rodents & others & training & 39 & 35,549\\
Sandy Related & NYC OpenData & training & 38 & 87,444 \\
SFR55\_2017\_national\_characteristics & gov.uk & test & 41 & 735\\
SFR55\_2017\_one\_plus\_sessions & gov.uk & test & 31 & 228,282\\
Survey & others & test & 27 & 1,259 \\
Tao & github.com & training & 7 & 736 \\
Tb & github.com  & training & 23 & 5,769 \\
TundraTraits & github.com & training & 17 & 73,428\\
User Profile                     & UCI ML              & training & 19 & 138\\
Vehicles 2015 & data.gov.uk & test & 23 & 257,845\\
4 csv files & mass.gov & test & 27 (avg.) & 46,934 (avg.)\\
9 csv files & data.gov & test & 14 (avg.) & 3904 (avg.)\\
2015ReportedTaserData & github.com & training & 69 & 610\\
16 csv files & T2Dv2 Gold Standard & test & 5 (avg.) & 127 (avg.) \\
83492acc-1aa2-4e80-ad05-28741e06e530.csv & github.com & training & 15 & 886\\
\bottomrule
\end{tabular}
}
\caption{Information about the data sets used.}
\label{table:datasets}
\end{table}

\clearpage

\begin{figure}[ht!]
  \centering
    \includegraphics[width=.7\textwidth]{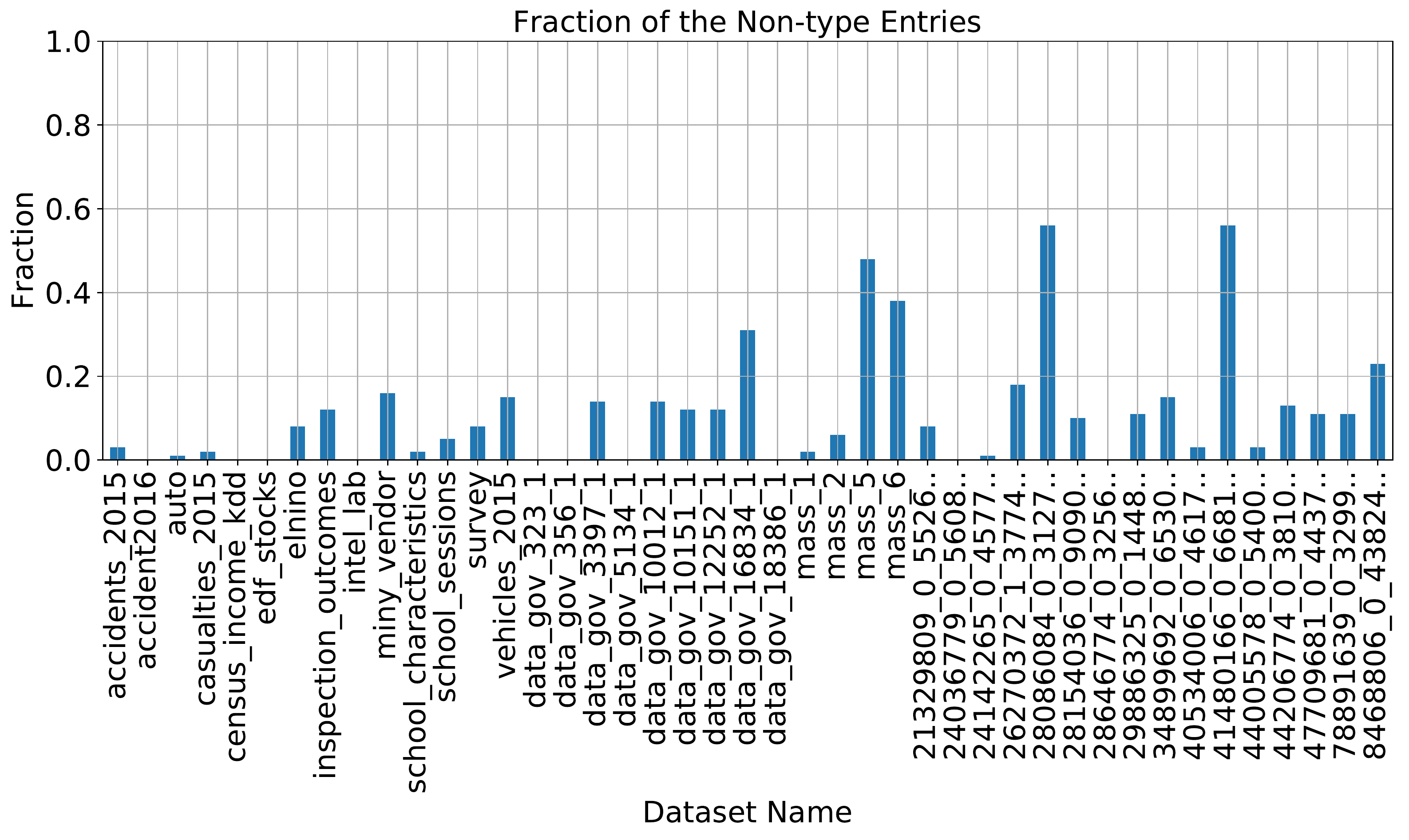}
  \caption{Fraction of the non-type entries in a dataset, calculated by aggregating over its columns. Note that `overall' denotes the fraction after aggregating over the datasets.}
\label{fig:nontype_fractions}
\end{figure}

\section{Derivations for the Training}
\label{sec:derivations_training}
The task is to update the parameters of the PFSMs, given a set of columns $X$ and their column types $\textbf{t}$. Since the columns are assumed to be independent, the gradient can be calculated by summing the gradient of each column. In the interest of simplicity, here we only derive the gradient for a column $\textbf{x}$ of type $k$. We would like to maximize the posterior probability of the correct type $k$ given a data column $\textbf{x}$, which can be rewritten as follows:
\begin{align}
\log p(t=k|\textbf{x}) &= \underbrace{\log p(t=k,\textbf{x})}_{L_c} - \underbrace{\log p(\textbf{x})}_{L_f}.
\end{align}
We now present the derivations of the gradients w.r.t.\ the transition parameters where $\theta^\tau_{q,\alpha,q'}$ denotes the transition parameter from state $q$ to $q'$ emitting the symbol $\alpha$ in the $\tau^{th}$ PFSM. Note that $\tau \in \{1,\dots,K\}$ where $K$ is the number of PFSMs.

We now differentiate these two terms, in section (4.1) and (4.2) respectively.
\subsection{Derivative of $L_c$} 
\begin{align}
\frac{\partial L_c}{\partial \theta_{q,\alpha,q'}^\tau} &= \frac{\partial \log p(t=k,\bf{x})}{\partial \theta_{q,\alpha,q'}^\tau}, \nonumber \\
&= \frac{\partial \Big( \log p(t=k) \prod_{i=1}^N p(x_i|t=k)\Big)}{\partial \theta_{q,\alpha,q'}^\tau}, \nonumber \\
&= \frac{\partial \Big( \log p(t=k) + \sum_{i=1}^N \log p(x_i|t=k)\Big)}{\partial \theta_{q,\alpha,q'}^\tau}, \nonumber \\
&= \sum_{i=1}^N \frac{\partial \log p(x_i|t=k)}{\partial \theta_{q,\alpha,q'}^\tau}, \nonumber \\
&= \sum_{i=1}^N \frac{1}{p(x_i|t=k)} \frac{\partial p(x_i|t=k)}{\partial \theta_{q,\alpha,q'}^\tau}, \nonumber \\
&= \sum_{i=1}^N \frac{1}{p(x_i|t=k)} \frac{\partial \big(\sum_{z'} p(z_i=z',x_i|t=k) \big)}{\partial \theta_{q,\alpha,q'}^\tau}, \nonumber \\
&= \sum_{i=1}^N \frac{1}{p(x_i|t=k)} \frac{\sum_{z'} \partial p(z_i=z',x_i|t=k)}{\partial \theta_{q,\alpha,q'}^\tau}, \nonumber \\
&= \sum_{i=1}^N \frac{1}{p(x_i|t=k)} \frac{\sum_{z'} \partial \Big(p(z_i=z'|t=k) p(x_i|z_i=z')\Big)}{\partial \theta_{q,\alpha,q'}^\tau}, \nonumber \\
&= \sum_{i=1}^N \frac{1}{p(x_i|t=k)} \frac{\partial \Big(\pi_k^k p(x_i|z_i=k) + \pi_k^m p(x_i|z_i=m) + \pi_k^a p(x_i|z_i=a) \Big)}{\partial \theta_{q,\alpha,q'}^\tau}, \nonumber \\
&= \sum_{i=1}^N \frac{1}{p(x_i|t=k)} \frac{\pi_k^k \partial p(x_i|z_i=k)}{\partial \theta_{q,\alpha,q'}^\tau}.
\end{align}

When $\tau$ is not equal to $k$, eq.\ (2) becomes 0. On the other hand, if $\tau=k$, then we would need to calculate $\frac{\partial p(x_i|z_i=\tau)}{\partial \theta_{q,\alpha,q'}^\tau}$, where $p(x_i|z_i=\tau)$ can be rewritten as $\sum_{q_{0:L}} p(x_i,q_{0:L}|z_i=\tau)$ as $x_i$ is generated by a PFSM. Note that $q_{0:L}$ denotes the states visited to generate $x_i$. The derivative can be derived as follows: 
\begin{align}
\frac{ \partial p(x_i|z_i=\tau)}{\partial \theta_{q,\alpha,q'}^\tau} &= \frac{ \partial \Big( \sum_{q_{0:L}} p(x_i,q_{0:L}|z_i=\tau) \Big)}{\partial \theta_{q,\alpha,q'}^\tau}, \nonumber \\
&= \sum_{q_{0:L}} \frac{\partial p(x_i,q_{0:L}|z_i=\tau)}{\partial \theta_{q,\alpha,q'}^\tau}, \nonumber \\
&= \sum_{q_{0:L}} p(x_i,q_{0:L}|z_i=\tau) \frac{\partial \log p(x_i,q_{0:L}|z_i=\tau)}{\partial \theta_{q,\alpha,q'}^\tau}, \nonumber \\
&= \sum_{q_{0:L}} p(x_i,q_{0:L}|z_i=\tau) \frac{\partial \log \Big[ I^{\tau}(q_0) \Big( \prod_{l=0}^{L-1} T^{\tau}(q_l, x_i^l, q_{l+1}) \Big) F^{\tau}(q_L)\Big]}{\partial \theta_{q,\alpha,q'}^\tau}, \nonumber \\
&= \sum_{q_{0:L}} p(x_i,q_{0:L}|z_i=\tau) \frac{\partial \sum_{l=0}^{L-1} \Big( \log T^{\tau}(q_l, x_i^l, q_{l+1}) \Big)}{\partial \theta_{q,\alpha,q'}^\tau}, \nonumber \\
&= \sum_{q_{0:L}} p(x_i,q_{0:L}|z_i=\tau) \sum_{l=0}^{L-1} \frac{\partial \log T^{\tau}(q_l, x_i^{l}, q_{l+1}) }{\partial \theta_{q,\alpha,q'}^\tau}, \nonumber \\
&= \sum_{q_{0:L}} p(x_i,q_{0:L}|z_i=\tau) \sum_{l=0}^{L-1} \frac{\delta(q_l,q) \delta(x_i^l, \alpha) \delta(q_{l+1}, q')}{T^{\tau}(q_l, x_i^l, q_{l+1})}, \nonumber \\
&= \sum_{q_{0:L}} \sum_{l=0}^{L-1} p(x_i,q_{0:L}|z_i=\tau) \Big(\frac{\delta(q_l,q) \delta(x_i^l, \alpha) \delta(q_{l+1}, q')}{T^{\tau}(q_l, x_i^l, q_{l+1})}\Big), \nonumber \\
&= \sum_{q_{0:L}} \sum_{l=0}^{L-1} p(q_l=q, q_{l+1}=q', q_{0:L \setminus l,l+1}, x_i|z_i=\tau) \Big(\frac{\delta(x_i^l, \alpha)}{T^{\tau}(q, x_i^l, q')}\Big), \nonumber \\
&= \sum_{l=0}^{L-1} \sum_{q_{0:L}} p(q_l=q, q_{l+1}=q', q_{0:L \setminus l,l+1}, x_i|z_i=\tau) \Big(\frac{\delta(x_i^l, \alpha)}{T^{\tau}(q, x_i^l, q')}\Big), \nonumber \\
&= \sum_{l=0}^{L-1}  
\frac{\delta(x_i^l, \alpha) p(q_l=q, q_{l+1}=q', x_i|z_i=\tau) }{T^{\tau}(q, x_i^l, q')}.
\end{align}
Hence, we need to evaluate the joint probability $p(q_l=q, q_{l+1}=q', x_i|z_i=\tau)$ for each $l$ where $x_i^l=\alpha$, which can be found by marginalizing out the variables $q_{0:L \setminus \{l,l+1\}}$:
\begin{align}
p(q_l=q, q_{l+1}=q', x_i|z_i=\tau) &= \sum_{q_{l'}} p(q_l=q, q_{l+1}=q', q_{l'}, x_i|z_i=\tau), 
\end{align}
where $l'$ denotes $\{0:L\} \setminus \{l,l+1\}$. This can be calculated iteratively via Forward-Backward Algorithm where the forward and backward messages are defined iteratively as follows:
\begin{align}
v_{l \rightarrow l+1}(q_l) &= \sum_{q_{l-1}} T^\tau(q_{l-1}, x_i^l,  q_l) v_{l-1 \rightarrow l}(q_{l-1}), \nonumber \\
\lambda_{l+1 \rightarrow l}(q_{l+1}) &= \sum_{q_{l+2}} T^\tau(q_{l+1}, x_i^{l+2},  q_{l+2}) \lambda_{l+2 \rightarrow l+1}(q_{l+2}), \nonumber \\
\end{align}

We can then rewrite $p(q_l=q, q_{l+1}=q', x_i|z_i=\tau)$ as follows:
\begin{align}
p(q_l, q_{l+1}, x_i|z_i=\tau) &= (v_{l \rightarrow l+1}(q_{l}) \bullet \lambda_{l+1 \rightarrow l}(q_{l+1})) \odot T^\tau(q_l, x_i^{l+1},  q_{l+1}), 
\end{align}
where $\bullet$ and $\odot$ denote respectively outer and element-wise product.

\subsection{Derivative of $L_f$}
Let us now take the derivative of the second term $L_f$:
\begin{align}
\frac{\partial L_f}{\partial \theta_{q,\alpha,q'}^\tau} &= \frac{\partial \log p(x)}{\partial \theta_{q,\alpha,q'}^\tau}, \nonumber \\
&= \frac{\partial \sum_{i=1}^N \log p(x_i)}{\partial \theta_{q,\alpha,q'}^\tau}, \nonumber \\
&= \sum_{i=1}^N \frac{\partial \log p(x_i)}{\partial \theta_{q,\alpha,q'}^\tau}, \nonumber \\
&= \sum_{i=1}^N \frac{1}{p(x_i)} \frac{\partial \big(\sum_{t'} \sum_{z'} p(t=t',z_i=z',x_i) \big)}{\partial \theta_{q,\alpha,q'}^\tau}, \nonumber \\
&= \sum_{i=1}^N \frac{1}{p(x_i)} \frac{\partial \big(\sum_{z'} p(t=\tau,z_i=z',x_i) \big)}{\partial \theta_{q,\alpha,q'}^\tau}, \nonumber \\
&= \sum_{i=1}^N \frac{1}{p(x_i)} \frac{\partial \big(\sum_{z'} p(t=\tau) p(z_i=z'|t=\tau) p(x_i|z_i=z') \big)}{\partial \theta_{q,\alpha,q'}^\tau}, \nonumber \\
&= \sum_{i=1}^N \frac{1}{p(x_i)} \frac{p(t=\tau) \partial \big(\pi_\tau^\tau p(x_i|z_i=\tau) + \pi_\tau^m p(x_i|z_i=m) + \pi_\tau^a p(x_i|z_i=a) \big)}{\partial \theta_{q,\alpha,q'}^\tau}, \nonumber \\
&= \sum_{i=1}^N \frac{1}{p(x_i)} \frac{p(t=\tau) \pi_\tau^\tau \partial p(x_i|z_i=\tau)}{\partial \theta_{q,\alpha,q'}^\tau}, \nonumber \\
&= \sum_{i=1}^N \frac{p(t=\tau) \pi_\tau^\tau}{p(x_i)} \frac{ \partial p(x_i|z_i=\tau)}{\partial \theta_{q,\alpha,q'}^\tau}.
\end{align}

Let us now put all the equations together. When we are calculating the derivative of eq.\ (1) w.r.t.\ the correct machine, i.e.\ $\tau=k$, the derivative becomes the following:
\begin{align}
\frac{\partial \log p(t=k|x)}{\partial \theta^\tau_{q,\alpha,q'}} &= \sum_{i=1}^N \Big( \frac{\pi_k^k}{p(x_i|t=k)} \frac{\partial p(x_i|z_i=k)}{\partial \theta_{q,\alpha,q'}^k} - \frac{\pi_k^k p(t=k)}{p(x_i)} \frac{ \partial p(x_i|z_i=k)}{\partial \theta_{q,\alpha,q'}^k} \Big), \nonumber \\
&= \sum_{i=1}^N \Big( \pi_k^k \frac{\partial p(x_i|z_i=k)}{\partial \theta_{q,\alpha,q'}^k} \Big( \frac{1}{p(x_i|t=k)} - \frac{p(t=k)}{p(x_i)}\Big) \Big), \nonumber \\
&= \sum_{i=1}^N \frac{\pi_k^k}{p(x_i|t=k)} \frac{\partial p(x_i|z_i=k)}{\partial \theta_{q,\alpha,q'}^k} \Big( 1 - \frac{p(t=k) p(x_i|t=k)}{p(x_i)}\Big), \nonumber \\
&= \sum_{i=1}^N \frac{\pi_k^k}{p(x_i|t=k)} \frac{\partial p(x_i|z_i=k)}{\partial \theta_{q,\alpha,q'}^k} \Big( 1 - \frac{p(t=k,x_i)}{\sum_{k'} p(t=k',x_i)}\Big).
\end{align}

When we are calculating the derivative of eq.\ (1) w.r.t.\ the wrong machines, i.e.\ $\tau \neq k$ this becomes:
\begin{align}
\frac{\partial \log p(t=k|x)}{\partial \theta^\tau_{q,\alpha,q'}} &= - \sum_{i=1}^N \Big( \frac{\pi_\tau^\tau p(t=\tau)}{p(x_i)} \frac{ \partial p(x_i|z_i=\tau)}{\partial \theta_{q,\alpha,q'}^\tau} \Big).
\end{align}

Lastly, we ensure the parameters remain positive and normalized using the softmax function. We define $T_\tau(q,\alpha,q') = \exp{T_\tau^z(q,\alpha,q')}/(\exp{F_\tau^z(q)} + \sum_{\alpha',q''} \exp{T_\tau^z(q,\alpha',q'')} )$ and $I_\tau^z(q) = \exp{I_\tau^z(q)}/\sum_{q'} \exp{I_\tau^z(q')}$. We now update these new unconstrained parameters using the new gradient calculated via the chain rule: $\partial f/\partial T_\tau^z(q,\alpha,q') = (\partial f/ \partial T_\tau(q,\alpha,q')) (\partial T_\tau(q,\alpha,q') /\partial T_\tau^z(q,\alpha,q'))$.

\section{Derivations for Inference}
\label{sec:derivations_inference}
The posterior distribution of column type $t$ can be derived as follows:
\begin{eqnarray*}
p(t=k|\textbf{x}) &\propto& p(t=k, \textbf{x}), \nonumber \\
&=& p(t=k) \prod_{i=1}^N p(x_i | t=k), \nonumber \\
&=& p(t=k) \prod_{i=1}^N \Big( \pi_k^k p(x_i | z_i = k) + \pi_k^m p(x_i | z_i = m) + \pi_k^a p(x_i | z_i = a) \Big).
\end{eqnarray*}
Let us assume that $t=k$ according to $p(t|\textbf{x})$, the posterior distribution of column type. Then we can write the posterior distribution of row type $z_i$ given $t=k$ and $\textbf{x}$ as:
\begin{eqnarray}
p(z_i=j | t=k, \textbf{x}) 
&=& \frac{p(z_i=j, x_i | t=k)}{p(x_i | t=k)}, \nonumber \\
&=& \frac{p(z_i=j, x_i | t=k) }{\sum_{z_i \in \{k,m,a\}} p(z_i, x_i | t=k)}.
\end{eqnarray}

\section{The Outputs of the PADS Library}
\label{sec:pads_appendix}
We have mentioned previously that the outputs generated by the PADS library do not directly address our problem. We present a sample from an example test dataset in Table \ref{table:442_sample}, and a part of the corresponding output of the PADS library.
\begin{table}[ht!]
\center
\scalebox{0.7}{
\large
\begin{tabular}{llllll}
\toprule
year    & winner/2nd        & NULL & scores         & total & money (us\$) \\
\midrule
1998    & fred couples      & 1    & 64-70-66-66-66 & 332   & 414000       \\
\&nbsp; & bruce lietzke     & 2    & 65-65-71-62-69 & 332   & 248400       \\
1997    & john cook         & 1    & 66-69-67-62-63 & 327   & 270000       \\
\&nbsp; & mark calcavecchia & 2    & 64-67-66-64-67 & 328   & 162000       \\
1996    & mark brooks       & 1    & 66-68-69-67-67 & 337   & 234000       \\
\&nbsp; & john huston       & 2    & 69-71-65-65-68 & 338   & 140400       \\
1995    & kenny perry       & 1    & 63-71-64-67-70 & 335   & 216000       \\
\&nbsp; & david duval       & 2    & 67-68-65-67-69 & 336   & 129600       \\
1994    & scott hoch        & 1    & 66-62-70-66-70 & 334   & 198000       \\
\&nbsp; & fuzzy zoeller     & t2   & 70-67-66-68-66 & 337   & 82133.34     \\
\&nbsp; & lennie clements   & t2   & 67-69-61-72-68 & 337   & 82133.33     \\
\&nbsp; & jim gallagher jr. & t2   & 66-67-74-62-68 & 337   & 82133.33     \\
1993    & tom kite          & 1    & 67-67-64-65-62 & 325   & 198000       \\
\bottomrule
\end{tabular}
}
\caption{A sample test dataset.}
\label{table:442_sample}
\end{table}

The outputs are interpreted starting from the bottom. In this case, the data is defined as an array of ``struct'' type named Struct\_194. This is further characterized as a combination of various ``union'' types. For example, let us consider the first one named Union\_19 which consists of a constant string $\texttt{\&nbsp;}$, another constant string $\texttt{year}$, and and integer type. However, this can be more complicated as in type Union\_165 consisting of two struct types Struct\_192 and Struct\_164. Note that the former is further divided into a union type, whereas the latter is described as a combination of some constant strings and a float type. As the reader can see, it can become difficult and time-consuming to interpret an output. Moreover, the output becomes more complex when delimiters are inferred correctly, as this can prevent the types from column specific.

\begin{figure}
\begin{lstlisting}
#include "vanilla.p"
Punion Union_19 {
	v_stringconst_12 Pfrom("&nbsp;");
	"year";
	Puint16  v_intrange_4;
};
Pstruct Struct_76 {
	"\"winner/";
	Puint8  v_intconst_70 : v_intconst_70 == 2;
	"nd\"";
};
.
.
.
Punion Union_189 {
	v_stringconst_173 Pfrom("money (us$)");
	Puint32  v_intconst_169 : v_intconst_169 == 72600;
};
Pstruct Struct_192 {
	'\"';
	Union_189  v_union_189;
	'\"';
};
Pstruct Struct_164 {
	'\"';
	Pfloat64  v_float_156;
	'\"';
};
Punion Union_165 {
	Struct_192  v_struct_192;
	Struct_164  v_struct_164;
};
Precord Pstruct Struct_194 {
	'\"';
	Union_19  v_union_19;
	"\",";
	Union_62  v_union_62;
	",\"";
	Union_86  v_union_86;
	"\",";
	Union_121  v_union_121;
	',';
	Union_141  v_union_141;
	',';
	Union_165  v_union_165;
};
Psource Parray entries_t {
	Struct_194[];
};

\end{lstlisting}
\caption{A fragment of the PADS output for a given dataset.}
\end{figure}

\clearpage

\section{Additional Experimental Results}
\label{sec:additional_exp_results}
Table \ref{table:evaluations_hypoparsr} presents the comparisons with hypoparsr.

\begin{table}[h!]
\centering
\scalebox{0.7}{
\large
\begin{tabular}{lllllllll}
\toprule 
\multicolumn{1}{c}{} &  \multicolumn{7}{c}{\bfseries Method} \\\cline{2-9} \addlinespace[1mm] 
  &   F\# & hypoparsr & messytables & readr &  TDDA& Trifacta & ptype-hc & ptype \\
\midrule
Overall  & \multirow{2}{*}{0.65} &  \multirow{2}{*}{0.66} &  \multirow{2}{*}{0.65} & \multirow{2}{*}{0.63} &   \multirow{2}{*}{0.60} & \multirow{2}{*}{0.88} & \multirow{2}{*}{\bfseries 0.96} & \multirow{2}{*}{0.95} \\
Accuracy &&&&&&& \\
\hline 
\addlinespace[1mm] 
Date    & 0.31 &  0.31 &  0.17 &  0.07 & 0.00 & 0.62 & \bfseries 0.66  & \bfseries 0.66 \\
Logical & 0.27 &  0.00 & 0.29 &  0.00 & 0.00 & 0.14 & \bfseries 0.88 & \bfseries 0.88  \\
Numeric  & 0.43 &  0.52 & 0.43 &  0.45 & 0.39 & 0.88 & \bfseries 0.94 & 0.93 \\
Text  & 0.56 &  0.54 & 0.57 & 0.55 & 0.52 & 0.80 & \bfseries 0.94 & 0.93\\
\bottomrule
\end{tabular}
}
\caption{Performance of the methods using the Jaccard index and overall accuracy, for the types Date, Logical, Numeric and Text.}
\label{table:evaluations_hypoparsr}
\end{table}

\begin{figure}[ht!]
	\begin{subfigure}[b]{0.33\textwidth}
    	\captionsetup{justification=centering}
		\includegraphics[width=.8\textwidth]{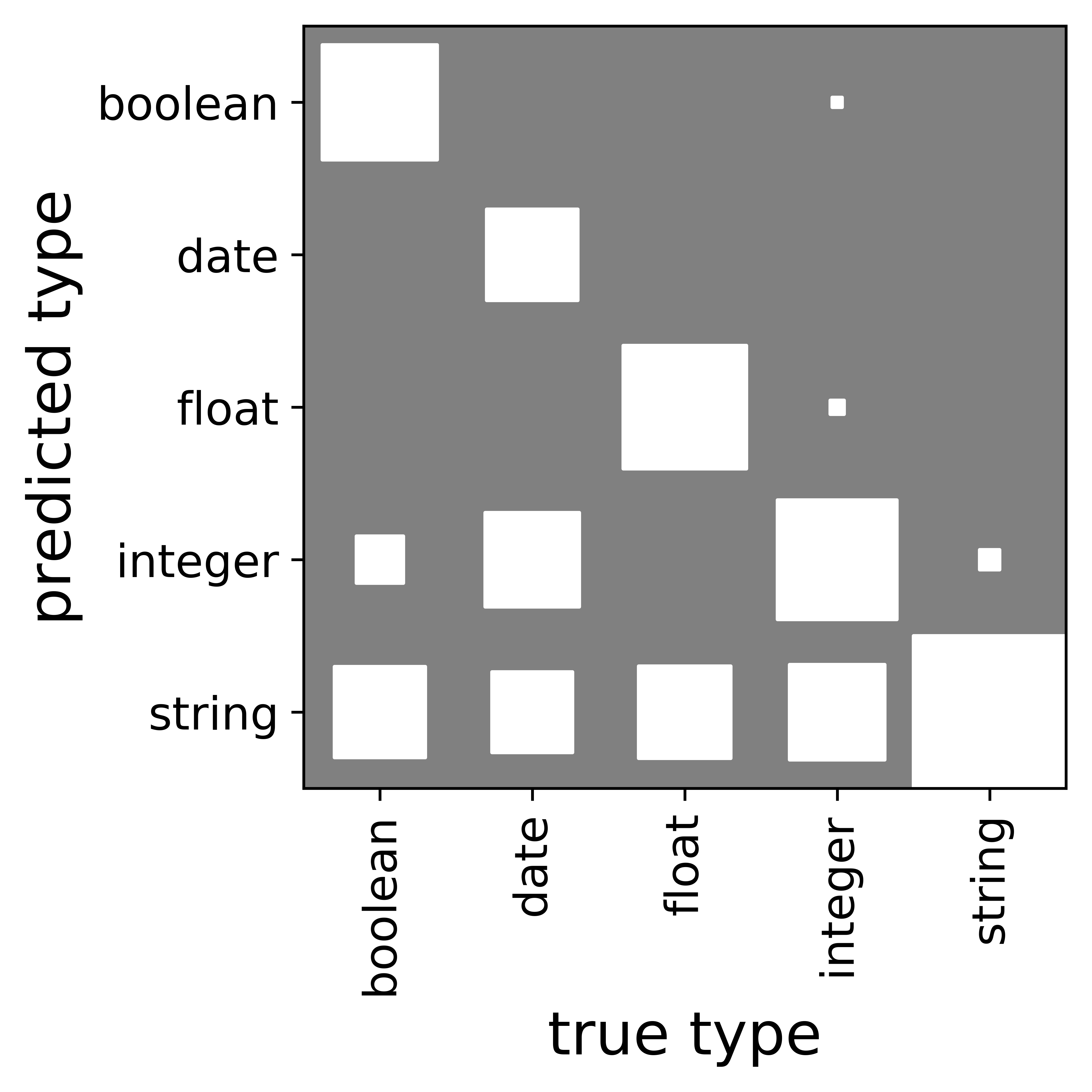}
		\caption{F\#}
	\end{subfigure}%
    \hfill
	\begin{subfigure}[b]{0.33\textwidth}    
    	\captionsetup{justification=centering}
		\includegraphics[width=.8\textwidth]{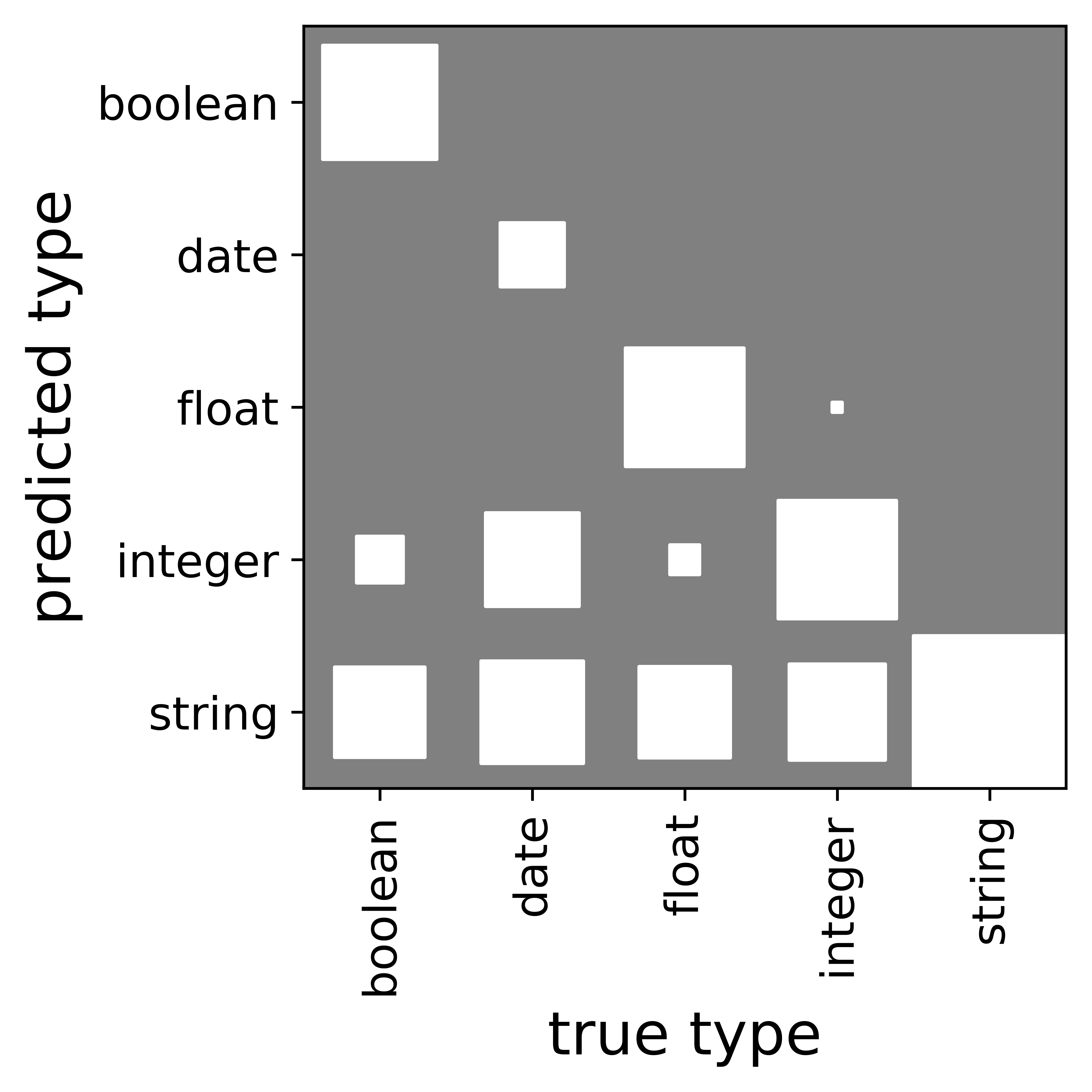}
		\caption{messytables}
	\end{subfigure}
	
	\begin{subfigure}[b]{0.33\textwidth}    
    	\captionsetup{justification=centering}
		\includegraphics[width=.8\textwidth]{Fig2a.png}
		\caption{ptype}
	\end{subfigure}%
    \hfill
    \begin{subfigure}[b]{0.33\textwidth}
		\captionsetup{justification=centering}
        \includegraphics[width=.8\textwidth]{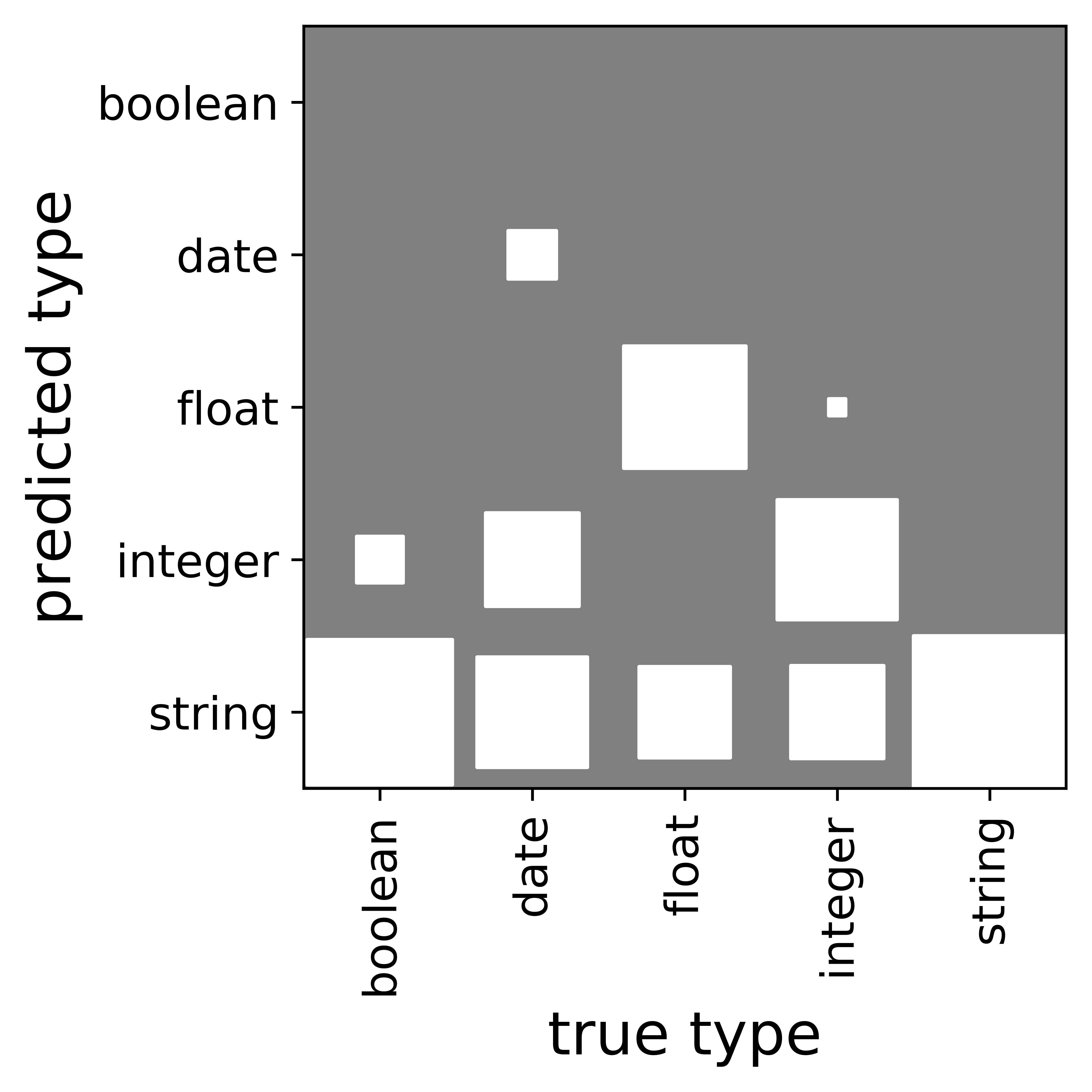}
		\caption{readr}
	\end{subfigure}
	
	\begin{subfigure}[b]{0.33\textwidth}
		\captionsetup{justification=centering}
        \includegraphics[width=.8\textwidth]{Fig2b.png}
		\caption{Trifacta}
	\end{subfigure}%
    \hfill
    \begin{subfigure}[b]{0.33\textwidth}    
		\captionsetup{justification=centering}
        \includegraphics[width=.8\textwidth]{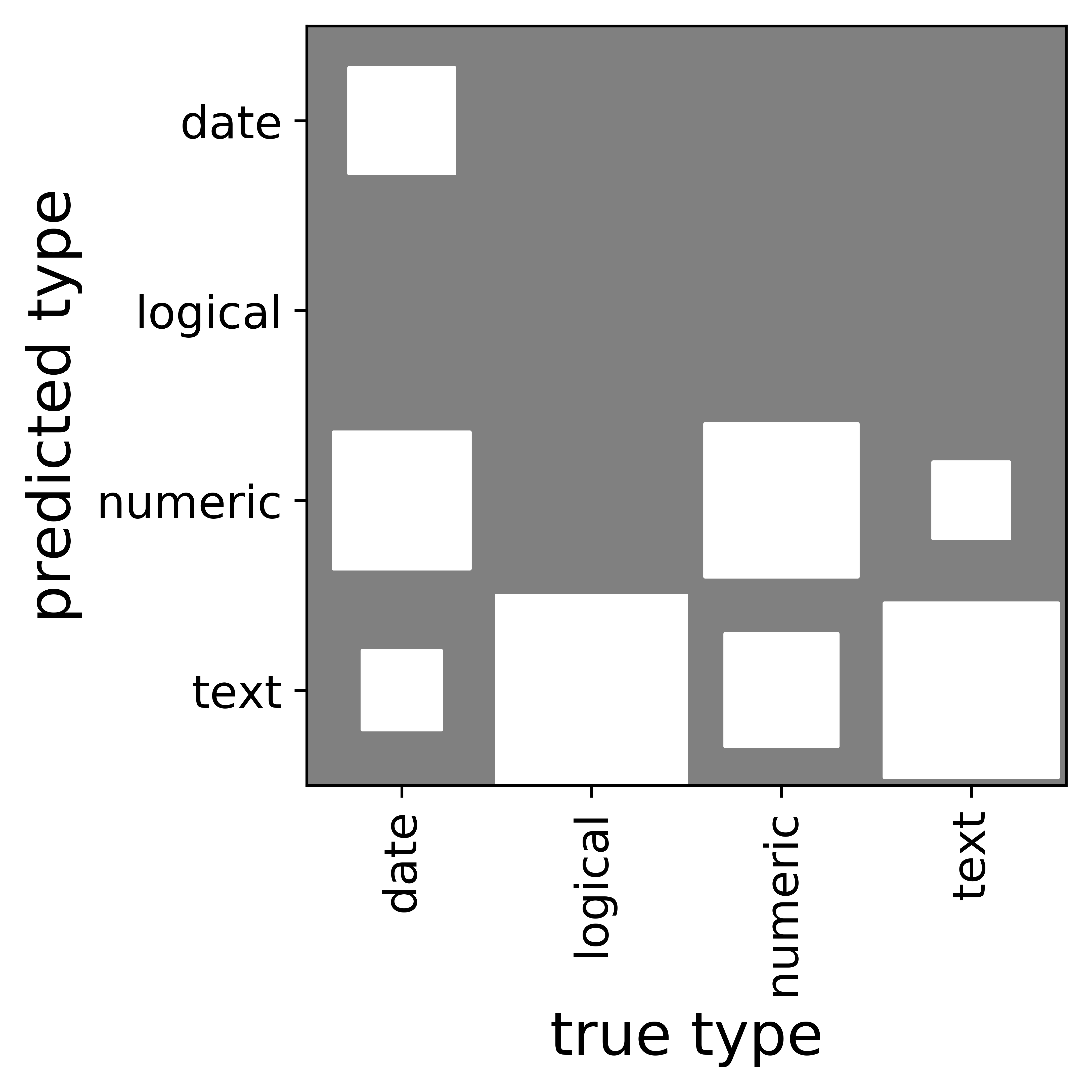}
		\caption{hypoparsr}
	\end{subfigure}	
    \caption{Normalized confusion matrices for (a) F\#, (b) messytables, (c) ptype, (d) reader, (e) Trifacta, and (f) hypoparsr plotted as Hinton diagrams, where the area of a square is proportional to the magnitude of the entry.}
    \label{fig:confusion_matrices_methods_appendix}
\end{figure}

Figure \ref{fig:confusion_matrices_methods_appendix} presents the normalized confusion matrices for the methods, discussed in the paper. 

\section{Scalability of the Methods}
\label{sec:scalability_appendix}
Table \ref{table:scalability_test} denotes the number of rows, columns, unique elements, and the time passed to infer column types.
\begin{table}[ht!]
\center
\scalebox{0.7}{
\large
\begin{tabular}{llllllllll}
\toprule
dataset &  \# cols &   \# rows &  $U$ & hypoparsr &  messytables &  ptype &  readr &  TDDA &Trifacta \\
\midrule
21329809\_0\_\dots&       6 &      156 &                    699 &         2.595 &            0.001 &      0.011 &      0.010 & 0.0001 & 1.333\\
 24036779\_0\_\dots&       7 &       83 &                    490 &         4.836 &            0.001 &      0.011 &      0.015 & 0.0001 & 1.143\\
 24142265\_0\_\dots&       6 &      100 &                    271 &         2.345 &            0.001 &      0.007 &      0.002 & 0.0001 & 1.333\\
 26270372\_1\_\dots&       5 &       19 &                     71 &         1.007 &            0.001 &      0.004 &      0.002 & 0.0001 & 2.200\\
 28086084\_0\_\dots&       6 &      224 &                    229 &         1.475 &            0.002 &      0.006 &      0.002 & 0.0001 & 1.333\\
  28154036\_0\_\dots&       5 &       10 &                     39 &         2.199 &            0.001 &      0.003 &      0.004 & 0.0001 & 1.400\\
 28646774\_0\_\dots&       6 &        8 &                     45 &         1.613 &            0.001 &      0.004 &      0.002 & 0.0001 & 1.333\\
 29886325\_0\_\dots&       6 &      254 &                   1063 &         3.572 &            0.002 &      0.015 &      0.014 & 0.0001 & 1.667\\
 34899692\_0\_\dots&       4 &       92 &                    321 &         0.978 &            0.001 &      0.009 &      0.001 & 0.00004 & 2.250\\
 40534006\_0\_\dots&       4 &       39 &                    118 &         2.667 &            0.001 &      0.004 &      0.002 & 0.00004 & 2.000\\
 41480166\_0\_\dots&       6 &      224 &                    229 &         1.343 &            0.001 &      0.005 &      0.002 & 0.0001 & 1.500\\
 44005578\_0\_\dots&       4 &        8 &                     31 &         2.014 &            0.001 &      0.003 &      0.001 & 0.00004 & 2.000\\
 44206774\_0\_\dots&       6 &       89 &                    279 &         2.505 &            0.031 &      0.006 &      0.002 & 0.0001& 1.500\\
 47709681\_0\_\dots&       4 &      408 &                    706 &         3.432 &            0.006 &      0.021 &      0.002 & 0.00004 & 2.250\\
 78891639\_0\_\dots&       6 &      202 &                    862 &         2.927 &            0.002 &      0.013 &      0.002 & 0.0001 & 1.333\\
  8468806\_0\_\dots&       7 &      110 &                    489 &         3.654 &            0.001 &      0.007 &      0.002 & 0.0001 & 1.143\\
                   accident2016 &      18 &      555 &                   1835 &        10.022 &            0.005 &      0.009 &      0.006 & 0.0001 & 0.611\\
                 accidents\_2015 &      32 &   140056 &                 609343 &             - &            0.895 &      1.635 &      0.045 & 0.0003 & 0.281\\
                           auto &      26 &      205 &                    911 &         3.557 &            0.001 &      0.011 &      0.006 & 0.0002 & 0.308\\
                casualties\_2015 &      16 &   186189 &                 140158 &             - &            1.191 &      0.858 &      0.030 & 0.0002 & 0.625\\
              census\_inc\dots &      42 &   199523 &                 102028 &             - &            0.853 &      0.423 &      0.054 & 0.0004 & 0.238\\
               data\_gov\_10012\_1 &      14 &      145 &                    779 &         4.023 &            0.003 &      0.012 &      0.011 & 0.0002 & 0.571\\
               data\_gov\_10151\_1 &      21 &       99 &                    775 &         2.227 &            0.002 &      0.006 &      0.002 & 0.0002 & 0.429\\
               data\_gov\_12252\_1 &       5 &      258 &                    756 &         2.734 &            0.002 &      0.015 &      0.002 & 0.00005 & 1.600\\
               data\_gov\_16834\_1 &      24 &    15055 &                   2911 &             - &            0.056 &      0.035 &      0.005 & 0.0002& 0.333\\
               data\_gov\_18386\_1 &      16 &     1238 &                   3975 &        17.705 &            0.007 &      0.024 &      0.002 & 0.0001& 0.500\\
                 data\_gov\_323\_1 &       6 &    13260 &                  11740 &        94.726 &            0.065 &      0.157 &      0.006 & 0.00005 & 1.500 \\
                data\_gov\_3397\_1 &      18 &      437 &                    534 &         5.785 &            0.002 &      0.005 &      0.002 & 0.0002 & 0.500\\
                 data\_gov\_356\_1 &       8 &     1279 &                   6993 &        25.147 &            0.008 &      0.068 &      0.003 & 0.0001& 1.000\\
                data\_gov\_5134\_1 &      10 &     3366 &                  12198 &       293.392 &            0.023 &      0.110 &      0.007 & 0.0001& 0.700\\
                     edf\_stocks &       7 &     5425 &                   4702 &        23.288 &            0.049 &      0.063 &      0.005 & 0.0001& 1.143\\
                         elnino &       9 &      782 &                   1032 &         12.59 &            0.005 &      0.012 &      0.002 & 0.0001 & 1.220\\
            inspection\_ou\dots &      22 &     1477 &                   3115 &        12.349 &            0.007 &      0.014 &      0.002 & 0.0002 & 0.364\\
                      intel\_lab &       8 &  1048576 &                 112912 &             - &            8.956 &      2.183 &      0.266 & 0.0001 & 1.250\\
                         mass\_1 &      12 &   131316 &                 803207 &             - &            0.703 &     12.502 &      0.150 & 0.0001& 0.750\\
                         mass\_2 &      19 &    44990 &                  12452 &             - &            0.275 &      0.137 &      0.015 & 0.0002 & 0.474\\
                         mass\_5 &      53 &     8282 &                  72659 &             - &            0.029 &      0.115 &      0.009 & 0.001 & 0.189\\
                         mass\_6 &      23 &     3148 &                   3363 &        36.681 &            0.012 &      0.018 &      0.002 & 0.0002 & 0.348\\
                    miny\_vendor &      18 &      897 &                   6728 &         7.148 &            0.006 &      0.030 &      0.002 & 0.0002 & 0.500\\
         school\_char\dots &      41 &      735 &                  16832 &         3.873 &            0.005 &      0.034 &      0.001 & 0.0003& 0.220\\
                school\_sessions &      31 &   228282 &                  78913 &             - &            1.010 &      0.514 &      0.039 & 0.0003 & 0.290\\
                         survey &      27 &     1259 &                   1622 &         7.036 &            0.008 &      0.015 &      0.001 & 0.0003& 0.296\\
                  vehicles\_2015 &      23 &   257845 &                 141278 &             - &            1.641 &      0.710 &      0.034 & 0.0002 & 0.391
\end{tabular}
}
\caption{Size of the test datasets and the times in seconds it takes to infer column types per column (on average), where $U$ denotes the number of unique data entries in a dataset.}
\label{table:scalability_test}
\end{table}

\normalsize{}
\end{document}